\documentclass{article}

\usepackage[preprint]{neurips_2026}

\usepackage{amsmath,amsfonts,bm}
\usepackage{booktabs}
\usepackage{multirow}
\usepackage{graphicx}
\usepackage{xcolor}
\usepackage{url}
\usepackage{amssymb}
\usepackage{wrapfig}
\usepackage{placeins}
\usepackage{algorithm}
\usepackage{algpseudocode}
\usepackage{float}
\usepackage{colortbl}
\usepackage{listings}   
\lstdefinestyle{prompt}{basicstyle=\scriptsize\ttfamily, breaklines=true, breakatwhitespace=true,
  columns=fullflexible, keepspaces=true, showstringspaces=false, frame=single, framerule=0.3pt,
  xleftmargin=3pt, xrightmargin=3pt, aboveskip=6pt, belowskip=6pt}
\usepackage{hyperref}
\definecolor{gtp}{RGB}{225,236,250}   
\definecolor{gtd}{RGB}{253,235,215}   

\title{SplatScout: Query-Driven Part-Level Referring Segmentation in Gaussian Scenes}

\author{%
  Yihan Zang \quad Da Li \quad Shinkyu Park \quad Ivan Viola \quad Dominik Engel \\
  CEMSE Division, King Abdullah University of Science and Technology \\
  \texttt{\{yihan.zang, da.li, shinkyu.park, ivan.viola, dominik.engel\}@kaust.edu.sa}
}

\newcommand{\dsname}{PartRef-ScanNet++}
\newcommand{\method}{SplatScout}

\newcommand{\rev}[1]{#1}

\begin{document}

\maketitle

\begin{abstract}

Existing referring segmentation methods on Gaussian
scenes primarily target whole objects, relying on
segmenter-generated candidates from which a
vision-language model (VLM) selects. \rev{In such a two-stage process, the candidates} may omit fine-grained parts needed
as targets or cues for identifying objects.
We introduce \method{}, a query-driven,
closed-loop approach that places perception
under VLM control. The model actively selects
views, directs segmentation, and inspects
rendered 3D results to refine its predictions.
For part-level evaluation, we introduce
\dsname{}, comprising 261 manually verified
queries across 12 ScanNet++ scenes with
mesh-derived ground truth.
On PartRef-ScanNet++, \method{} achieves
0.538 mIoU, substantially outperforming
existing methods. Ablations demonstrate
the benefits of iterative refinement,
while comparisons across VLM backbones
\rev{show that \method{} makes fuller use of
stronger VLMs}.
\method{} also achieves higher scores
on four existing referring benchmarks.
Code and the \dsname{} benchmark are available at
\url{https://github.com/kaust-violab/SplatScout}.
\end{abstract}

\section{Introduction}

3D Gaussian Splatting~\citep{kerbl20233d} represents scenes as explicit primitives that can be reconstructed and rendered in real time. Beyond rendering, a growing body of work attaches language to Gaussian scenes. Open-vocabulary methods distil features from 2D foundation models into Gaussian primitives, enabling category-level queries~\citep{langsplat,legaussians,opengaussian}. Referring methods go further, identifying specific objects described by expressions involving attributes and spatial relations~\citep{refersplat,zerosplat,causalsplat,gaussdet}. However, existing referring methods for Gaussian scenes primarily focus on whole objects, while practical instructions often refer to their parts, such as ``the handle of the second drawer on the right of the table by the window''. Unlike whole objects, parts have query-dependent granularity: the same chair may be requested as a whole in one query, its backrest in another, and its armrest in a third. Parts may also serve as targets, discriminators between similar objects, or anchors for locating other objects.

Grounding such expressions requires identifying the relevant scene regions and segmenting them at the granularity specified by the query. Suitable observations must be selected to obtain segmentation masks across multiple views, which are then integrated to construct the target Gaussian instance. However, the required views and segmentation regions depend on the query and may only become apparent during perception. A fixed, query-independent scene decomposition therefore cannot reliably provide all the candidates needed for grounding.

Meanwhile, modern vision-language models (VLMs) are increasingly capable of recognizing visual attributes, distinguishing fine-grained regions, and reasoning about spatial relationships in 2D images. Yet many existing grounding methods
\rev{follow a two-stage process that limits}
the VLM to selecting candidates proposed by a segmenter, either during preprocessing
\unskip\rev{~\citep{gaussdet,causalsplat,zsvg3d,csvg,seeground,seqvlm}}
or in a single pass at query time
\unskip\rev{~\citep{zerosplat,vlmgrounder}}
\unskip. Even when candidates are generated on demand,
\rev{as in OpenGround~\citep{openground} and Think, Act, Build (TAB)~\citep{tab},}
the VLM typically cannot use intermediate visual feedback to revise them. Consequently, if the target part is missing or segmented at the wrong granularity, stronger visual reasoning alone cannot recover it through candidate selection. This design limits how effectively advances in VLM capabilities translate into better grounding.

We address this limitation by
\rev{breaking away from this two-stage process and}
changing the role of the VLM from candidate selection to perception control. Rather than choosing from a predetermined set of instances, the model decides where to look, what to segment, and at what granularity, based on the expression and its observations. The segmenter and geometric operations become tools that execute these decisions, allowing the VLM to apply its visual understanding directly to the construction of query-specific 3D instances.

However, query-time perception alone is insufficient if decisions are executed in a single pass. The appropriate next action often depends on previous results: an observation may require closer inspection, segmentation may reveal an incorrectly selected region, and lifting the masks into 3D may expose errors that were not apparent in the original images. We therefore organize perception as a closed loop in which each operation returns feedback to the VLM. By inspecting intermediate results, the model can revise its decisions and progressively refine the target rather than committing to its initial predictions.

On this basis, we introduce \method{}, a query-driven, closed-loop perception system for
\rev{part-level}
referring segmentation in Gaussian scenes. Before a query arrives, \method{} never constructs semantic candidates. During query execution, the VLM coordinates three
\rev{tool sets}
\unskip: the Navigator selects informative observations, the Segmenter produces query-directed masks, and the Lifter constructs 3D Gaussian instances. Each tool returns feedback that informs subsequent decisions, including rendered views of lifted instances for detecting and correcting errors. The VLM thus controls both instance construction and the decision of when to return an answer.
\rev{To our knowledge, \method{} is the first zero-shot method for part-level referring segmentation in 3D scenes in which the VLM is not bound to a fixed procedure.}

To evaluate part-level grounding, we introduce \dsname{}, a benchmark containing 261 manually verified referring expressions across 12 ScanNet++\citep{yeshwanth2023scannet++} scenes. It covers whole objects, parts serving different roles in referring expressions, and multiple-target queries, with ground truth verified on laser-scanned meshes. On this benchmark, \method{} substantially outperforms prior referring segmentation methods and achieves higher scores on four of five existing referring benchmarks. Controlled experiments further investigate the benefits of closed-loop perception and the role of VLM capabilities in utilizing perceptual feedback.

Our main contributions are:
\begin{itemize}

\item We identify and experimentally investigate the limitations of
\rev{two-stage candidate-selection}
pipelines and compare them to a VLM using visual feedback to guide perception, particularly for part-level referring expressions.

\item We introduce \method{}, which transfers perceptual decision-making to the VLM and coordinates observation, segmentation, and 3D instance construction within a query-driven, closed-loop framework, without precomputed semantic candidates.

\item We introduce \dsname{}, a part-level referring segmentation benchmark with 261 human-verified expressions across 12 ScanNet++ scenes and an evaluation protocol based on mesh ground truth.

\item \method{} substantially improves referring segmentation performance on existing object-level benchmarks and the part-level tasks in \dsname{}. Controlled experiments demonstrate the benefits of using perceptual feedback to guide grounding
\unskip\rev{, and show that \method{} makes fuller use of stronger VLMs than two-stage methods}
\unskip.

\end{itemize}

\section{Related Work}

\paragraph{\rev{Language-driven segmentation in Gaussian scenes:}}
Many works attach language to
\rev{Gaussian}
primitives, primarily enabling open-vocabulary segmentation through object names or category labels rather than full referring expressions. They train features from 2D foundation models into per-scene fields~\citep{langsplat,legaussians,feature3dgs}, attach semantic information to Gaussian groups~\citep{gaussiangrouping,opengaussian,instancegaussian,relags}, or back-project features without training~\citep{drsplat,ludvig,extrinsplat,lbg,normlift}. ReferSplat~\citep{refersplat} extends language-based Gaussian segmentation to full referring expressions by training a referring field per scene
\unskip\rev{, and later methods do so in a zero-shot manner with a VLM~\citep{gaussdet,causalsplat,zerosplat,realm}.}

\paragraph{\rev{Zero-shot 3D visual grounding:}}
\rev{Zero-shot methods avoid supervised training~\citep{scanrefer,referit3d} by bringing}
in a vision-language model at query time.
\rev{Earlier methods select the target from the proposals of an open-vocabulary 3D grounder~\citep{llmgrounder} or from}
a lookup table of segmented objects
\rev{built once per scene~\citep{zsvg3d,csvg,seeground,seqvlm,agentgrounder}. Others obtain candidates at query time: VLM-Grounder~\citep{vlmgrounder} detects them in the images, OpenGround~\citep{openground} extends the table online for targets it lacks, such as parts, and Think, Act, Build (TAB)~\citep{tab} lets an agent segment the target category and reconstruct the selected instance in 3D. All of them, including the zero-shot Gaussian methods above, share a two-stage process that restricts the VLM to}
choosing among candidates that a segmenter has produced.
In this work we focus on
\rev{lifting this restriction, letting the VLM decide where to look, what to segment, and what to redo.}

\paragraph{Part-level grounding:}
Existing part-aware Gaussian methods such as LangSplat and its successors~\citep{langsplat,relags,lbg} primarily rely on SAM and CLIP models to create pre-generated whole-object, part, and sub-part masks and language embeddings. Their supported granularity is therefore limited by the regions identified during preprocessing. On point clouds, part-level understanding has been studied mainly through functional elements
\unskip\rev{~\citep{scenefun3d}, grounded}
using 2D detectors and segmenters~\citep{fun3du,affordbot} or fixed superpoints~\citep{3dpu}
\unskip\rev{, and through articulated parts~\citep{openground}.}
Anywhere3D~\citep{anywhere3d} evaluates bounding-box grounding across granularities, while Articulate3D and MultiScan~\citep{articulate3d,multiscan} provide detailed part annotations. Existing benchmarks primarily treat parts as predefined targets, with limited coverage of their roles in identifying other objects. In contrast, \dsname{} includes parts as targets, discriminators, and relational anchors, with mesh-level segmentation ground truth.

\section{Method}

\subsection{Preliminaries}
\label{sec:prelim}

\paragraph{2D Gaussian Splatting:}
We represent each scene with 2D Gaussian
Splatting (2DGS)~\citep{huang20242d} as a set
$\mathcal{G}=\{g_j\}_{j=1}^{N}$ of $N$
planar disks. Each has a center,
two tangent axes, an opacity $\alpha_j$,
and a colour $c_j$. Compared with 3DGS,
it provides a surface-oriented
representation for segmentation.

For a frame $f$ with known camera pose and
$P$ pixels, let $p$ denote a pixel.
Ray-splat intersection orders the disks
by depth along each pixel ray, yielding
the compositing weight
\begingroup
\setlength{\belowdisplayskip}{2pt}
\setlength{\belowdisplayshortskip}{2pt}
\begin{equation}
w^{f}_{j}(p)
=\alpha_j G_j(p)
\prod_{k<j}\bigl(1-\alpha_k G_k(p)\bigr),
\label{eq:weight}
\end{equation}
\endgroup
\noindent where $G_j(p)$ is the Gaussian
falloff at the ray-disk intersection,
and $k<j$ indexes disks preceding $g_j$
along the ray. The rendered colour is
$C^{f}(p)=\sum_j w^{f}_{j}(p)c_j$.

The same weights render arbitrary
per-Gaussian attributes. Let
$a\in\mathbb{R}^{N\times d}$ contain
$d$-dimensional attributes for all
$N$ Gaussians. Their rendered values
are $A^{f}=W^{f}a$, where
$W^{f}\in\mathbb{R}^{P\times N}$ is the
sparse contribution matrix with
$W^{f}_{p,j}=w^{f}_{j}(p)$.
For a Gaussian subset
$\mathcal{S}\subseteq\mathcal{G}$,
its accumulated contribution at pixel
$p$ is
$\sigma^{f}_{\mathcal{S}}(p)
=\sum_{g_j\in\mathcal{S}}w^{f}_{j}(p)$.
Thresholding this contribution yields
the mask
$\{p:\sigma^{f}_{\mathcal{S}}(p)>\tfrac12\}$,
with occlusion handled by compositing.

\paragraph{Notation:}
We write $s$ for a group (masks of one thing in one or more frames, not yet lifted), $o$ for an object (groups the model has declared to be the same thing), and $e$ for an instance (a subset of $\mathcal{G}$ obtained by lifting). Only lifted instances can be returned as answers.

\subsection{Method Overview}
\label{sec:overview}

\begin{figure}[t]
\vspace{-2mm}
\centering
\includegraphics[width=\linewidth]{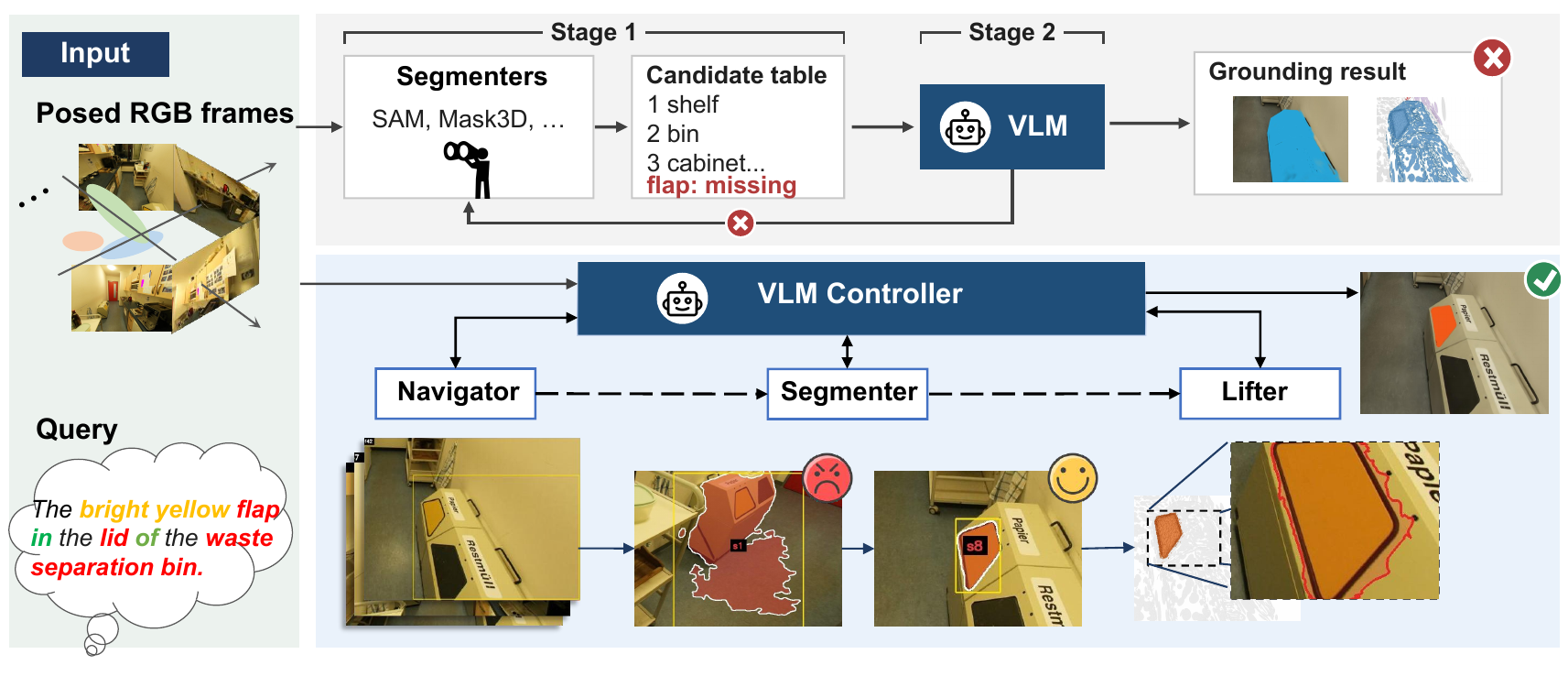}
\vspace{-20pt}
\caption{
Overview of \method{}.
\textbf{Top:} Existing \rev{two-stage} methods restrict the VLM
to selecting from segmentation candidates,
which may omit the referred target.
\textbf{Bottom:} \method{} enables query-driven,
closed-loop perception. The VLM coordinates
the Navigator, Segmenter, and Lifter to explore
views, extract target regions, and construct
3D-lifted Gaussian instances. Visual feedback allows
the VLM to inspect and refine intermediate
results without fixed semantic candidates.
}
\label{fig:overview}
\end{figure}

As illustrated in Fig.~\ref{fig:overview}, \method{}
consists of three functional tool sets coordinated by a VLM
controller: the Navigator, Segmenter, and Lifter. Starting
from the desired output, obtaining the target Gaussian subset
requires constructing 3D instances from multi-view segmentation
masks, which in turn requires acquiring suitable observations
of the scene. These dependencies define the responsibilities
of the three tool sets. The Navigator retrieves informative
scene views, the Segmenter produces query-directed masks,
and the Lifter integrates multi-view masks into 3D Gaussian
instances. Their complete interfaces, including available
actions, arguments, and feedback, are listed in
Table~\ref{tab:tools} of Appendix~\ref{app:loop}.

Rather than executing these tools in a predetermined sequence,
the VLM selects its next step action based on the referring
expression and the results of previous steps. Each tool
returns task-specific feedback, including scene observations,
segmentation overlays, or rendered views of lifted instances.
The VLM inspects these intermediate results and may invoke
additional tools to refine its current hypothesis.
Consequently, the target instances are constructed
progressively during query execution, without precomputed
semantic candidates.

\subsection{Closed-Loop Grounding}
\label{sec:loop}

Given a referring expression $Q$, let $E_t$
denote the intermediate entities constructed
at step $t$, including segmentation groups and
lifted 3D instances. Let $\mathcal{F}$ be the
available input frames, and
$\mathcal{H}_t\subseteq\mathcal{F}$ the frames
already observed by the VLM.
The interaction history is
$\tau_t=((a_0,r_0),\ldots,(a_{t-1},r_{t-1}))$,
where $a_i$ and $r_i$ denote an action and
its corresponding feedback.
Initially, $E_0$, $\mathcal{H}_0$, and
$\tau_0$ are empty.

At each step, the VLM selects an action
$a_t=\pi(Q,\tau_t,\ell_t)$, where $\pi$
denotes the VLM and $\ell_t$ summarizes
the execution state and remaining budget.
The tool interface $\Phi$ takes the action,
current entities, observed frames, Gaussian
scene, and available frames as inputs,
returning feedback $r_t$ and updated states
$E_{t+1}$ and $\mathcal{H}_{t+1}$.
The interaction history is then updated as
$\tau_{t+1}=\tau_t\Vert(a_t,r_t)$.

The tools execute operations but do not
decide whether to accept their results.
Based on the feedback, the VLM may explore
additional views, revise segmentations,
merge observations, or lift and inspect
3D instances. The execution order thus
adapts to the query and previous observations
rather than following a fixed pipeline.

The VLM terminates by returning a valid
answer, reporting that the target cannot
be identified, or exhausting its budget \rev{(16 steps or 150k input tokens)}.
A valid answer selects previously lifted
instances supported by multiple observations.
For the selected instance set $I$, the
final prediction is
$\mathcal{S}=\bigcup_{e\in I}\mathcal{G}_e$,
where $\mathcal{G}_e\subseteq\mathcal{G}$
is the Gaussian subset associated with
instance $e$. This also supports
multi-target queries.
The full grounding algorithm is provided
in Appendix~\ref{app:loop}.

\begin{figure}[t]
\vspace{-2mm}
\centering
\includegraphics[width=\linewidth]{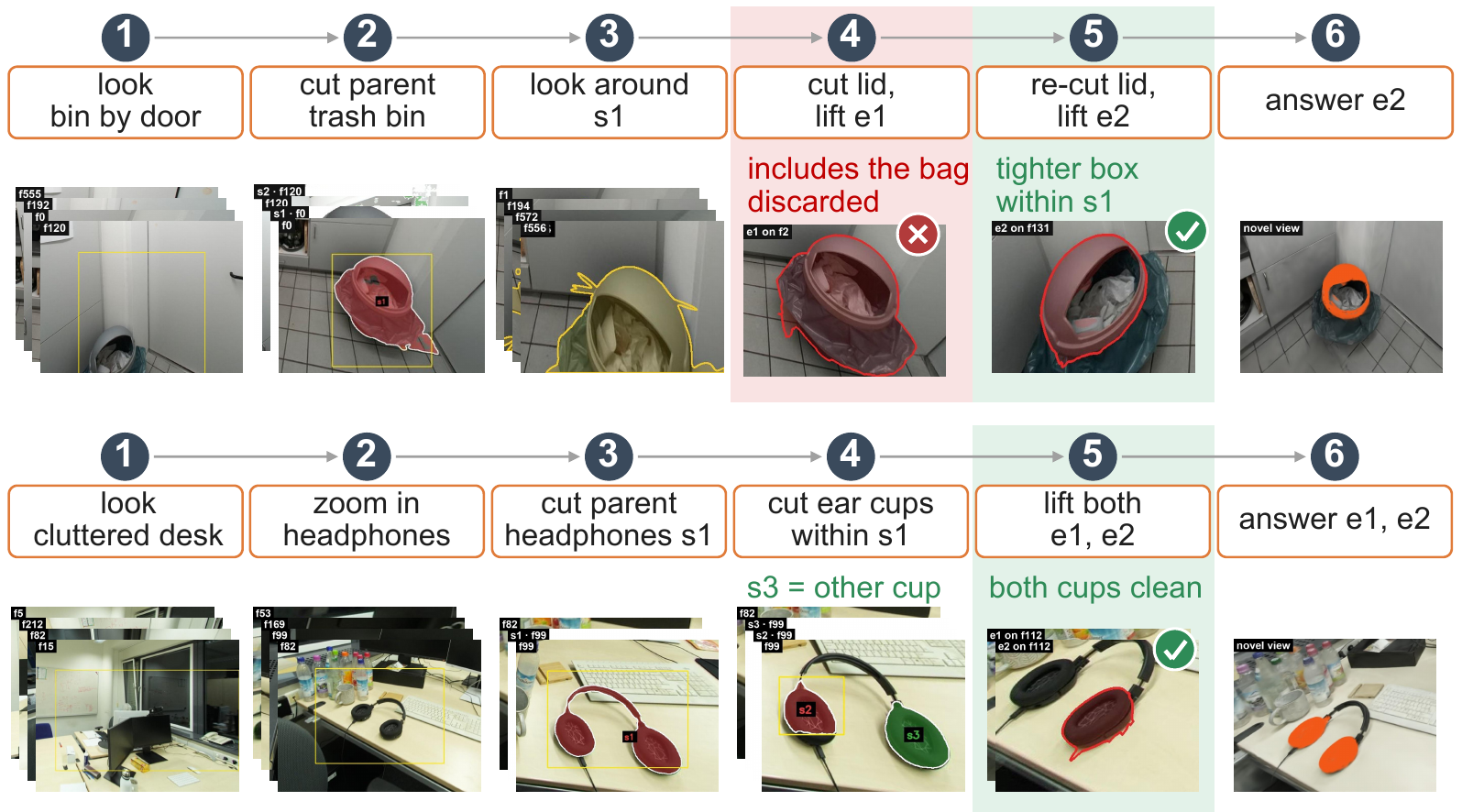}
\vspace{-14pt}
\caption{
Examples of closed-loop grounding.
\textbf{Top:} For ``the lid of the trash bin
beside the room door,'' render-back reveals
that the initial lift $e_1$ includes the bag.
The VLM re-segments within parent $s_1$
and accepts the corrected lift $e_2$.
\textbf{Bottom:} For ``the padded ear cups
of the black headphones on the desk,''
the VLM identifies and segments both ear cups
within the parent headphones, then lifts
and returns both instances $e_1$ and $e_2$.
}
\vspace{-16pt}
\label{fig:walkthrough}
\end{figure}

\subsection{Perception Tools}
\label{sec:tools}

The three tool sets in Figure.~\ref{fig:overview} provide the
perceptual and geometric operations used by the VLM
controller. The Navigator retrieves informative scene
observations, the Segmenter constructs query-directed
2D masks, and the Lifter converts these masks into
subsets of 2D Gaussian primitives in 3D space. Each tool returns intermediate
results that the VLM can inspect before selecting
its next action.

\subsubsection{Navigator: Active Scene Exploration}
\label{sec:navigator}

The Navigator provides two complementary capabilities:
global scene exploration and local inspection. Global
exploration helps the VLM locate relevant
regions, whereas local inspection provides
higher-resolution observations and complementary
viewpoints for examining a region.

\paragraph{Geometry-based scene preparation:}
To support efficient exploration without precomputing
semantic candidates, we prepare a lightweight geometric
index for each scene. For every training frame $f$,
we record the set of visible Gaussians $V_f$, together
with an approximate area weight $w_j$ for each Gaussian.
For any Gaussian subset
$\mathcal A\subseteq\mathcal G$,
we define its weighted area as
$w(\mathcal A)=\sum_{j:g_j\in\mathcal A}w_j$.

Given the set of previously observed frames
$\mathcal{H}$, their accumulated scene coverage is
\begin{equation}
    \operatorname{cov}(\mathcal{H})
    =
    \frac{
        w\left(\bigcup_{f\in\mathcal{H}}V_f\right)
    }{
        w(\mathcal{G})
    }.
    \label{eq:cov}
\end{equation}
This provides a geometry-based measure of how much
of the reconstructed scene has been observed, without
requiring semantic information.

\paragraph{Global exploration:}
The Navigator supports global exploration through
\texttt{explore}, which greedily selects frames
to increase scene coverage. Given the previously
observed frames $\mathcal{H}$, the next frame is
selected by
\begin{equation}
    f^{\star}
    =
    \arg\max_{f\notin\mathcal{H}}
    w\left(
        V_f \setminus
        \bigcup_{f'\in\mathcal{H}}V_{f'}
    \right).
    \label{eq:gain}
\end{equation}
Intuitively, each step selects the frame
that reveals the largest weighted portion
of the scene not yet observed.
Repeating this process yields complementary
views while avoiding redundant coverage.
The selected frames are returned as thumbnails
with their accumulated coverage, allowing
the VLM to identify promising regions or
request further exploration.

\paragraph{Local inspection:}
Once a potentially relevant region has been identified,
the Navigator supports closer inspection through
\texttt{nearby}. Given a reference frame, it retrieves
additional frames with moderate visual overlap, avoiding
both unrelated regions and nearly identical viewpoints.

For a specific region, the VLM may additionally provide
a bounding box $b$ in frame $f$. The Navigator returns
a high-resolution crop of the selected region and
retrieves other viewpoints observing approximately
the same reconstructed surface.

Specifically, let $\mathcal{G}_b$ denote the
visible Gaussians whose projected centers
lie inside the selected box $b$.
Candidate frames are identified according
to their visibility of $\mathcal{G}_b$.
To obtain complementary observations,
the Navigator selects viewpoints that
maximize angular diversity:
\begin{equation}
    f^{\star}
    =
    \arg\max_{f'\in\mathcal{K}\setminus\mathcal{P}}
    \min_{f''\in\mathcal{P}}
    \angle\left(
        \mathbf{d}_{f'},\mathbf{d}_{f''}
    \right),
    \label{eq:fps}
\end{equation}
where $\mathcal{K}$ denotes the candidate
frames and $\mathcal{P}$ contains the
already selected frames. Here, $f'$ is
a candidate frame, $f''$ is a previously
selected frame, and $\mathbf{d}_f$ denotes
the viewing direction from the selected
region toward camera $f$.
Selection is initialized with the frame
providing the greatest visibility of
$\mathcal{G}_b$.

The same strategy allows the VLM to revisit an
existing segmentation group or lifted instance
from previously unseen viewpoints. In this case,
the Navigator returns cropped images with the
corresponding region outlined, enabling visual
inspection before further segmentation or lifting.

\subsubsection{Segmenter: Query-Adaptive Segmentation}
\label{sec:segmenter}

\paragraph{VLM-directed segmentation:}
The Segmenter uses SAM3~\citep{sam3} to generate
query-specific masks from the frame, bounding box,
and text prompt selected by the VLM.
The VLM may also specify additional frames for
multi-view tracking. If joint text-and-box prompting
fails, the tool retries with the box alone.
The resulting masks are organized into segmentation
groups for subsequent 3D lifting.

\paragraph{Parent constraints and feedback:}
For part-level queries, the VLM may restrict
segmentation to an existing parent object.
The Segmenter projects the parent's Gaussian
representation into the selected frame and
suppresses regions outside its dilated mask
$m_{\mathrm{par}}$. A predicted mask $m$ is
retained if
$|m\cap m_{\mathrm{par}}|/|m|\geq 0.8$.

The Segmenter returns mask overlays, cut-outs,
multi-view results, and quality indicators,
including box-mask agreement and 3D overlap
with existing groups. These allow the VLM
to inspect results and decide
whether to revise them, explore
views, or proceed to lifting.

\subsubsection{Lifter: Multi-View Gaussian Instance Construction}
\label{sec:lifter}

The Lifter converts query-specific 2D masks
into 2D Gaussian instances in 3D space. Given segmentation
groups from one or more views, it aggregates
their geometric evidence, identifies the
Gaussian primitives, and returns
a rendered view of the instance
for inspection.

\paragraph{Multi-view mask aggregation:}
Before lifting, the Lifter organizes the masks
returned by the Segmenter into multi-view groups.
Each segmentation call produces a mask track
spanning the prompted frame and additional
tracked frames. Tracks are split at tracking
gaps to avoid incorrect associations.
The masks are then approximately back-projected
onto the Gaussians, and tracks with sufficient
3D overlap are merged into groups.
The VLM can further merge groups belonging
to the same physical entity using the
\texttt{same} tool. Each resulting group provides
the multi-view masks used for Gaussian lifting.

\paragraph{Gaussian lifting:}
We recover the Gaussian primitives corresponding
to each segmentation group using the
contribution-based voting formulation of
FlashSplat~\citep{flashsplat}.
For each group, we aggregate the multi-view
masks obtained through our closed-loop
perception process.

For a group $s$ with masks
$\{m^f\}_{f\in\mathcal F_s}$,
where $m^f\in\{0,1\}^{P}$, we accumulate
the foreground and background contributions
of all Gaussians:
\begin{equation}
    [\nu^1,\nu^0]
    =
    \sum_{f\in\mathcal F_s}
    (W^f)^\top
    [m^f,\mathbf{1}-m^f].
    \label{eq:vote}
\end{equation}
Here, $W^f\in\mathbb R^{P\times N}$ is
the rendering contribution matrix, and
$\nu^1$ and $\nu^0$ denote the accumulated
foreground and background contributions.
A Gaussian is assigned to the instance
when its foreground contribution exceeds
its background contribution:
\begin{equation}
    \mathcal{G}_e
    =
    \left\{
        g_j \in \mathcal{G} :
        \nu_j^1+\nu_j^0>0,\;
        \nu_j^1>\nu_j^0
    \right\}.
    \label{eq:majority}
\end{equation}
To improve multi-view consistency, when
multiple voting frames are available,
a selected Gaussian must additionally project
inside the segmentation masks of at least
two views. Very small Gaussian subsets
are discarded as fragments.

\paragraph{Iterative lifting:}
An initial lift uses the masks obtained from
the views selected during segmentation.
However, these observations may not capture
the complete geometry of the target.
When requested by the VLM, the Lifter performs
additional refinement rounds. At each round,
it retrieves complementary views in which
the current Gaussian instance is sufficiently
visible. The instance is projected into
these views to guide additional segmentation,
and the resulting masks are selected according
to their agreement with the rendered
instance silhouette.
New observations contribute additional
foreground and background votes, and the
Gaussian instance is updated using
Eq.~(\ref{eq:majority}).

Let $\mathcal{G}_e^{(k)}$ denote the Gaussian
subset of instance $e$ after refinement
round $k$, and let $|\cdot|$ denote
set cardinality. Refinement stops when
the relative change in the number of
selected Gaussians falls below a threshold:
\begin{equation}
    \frac{
        \left|
            |\mathcal{G}_e^{(k+1)}|
            -
            |\mathcal{G}_e^{(k)}|
        \right|
    }{
        \max\left(
            |\mathcal{G}_e^{(k)}|,1
        \right)
    }
    \leq 0.05,
    \label{eq:saturate}
\end{equation}
or when no suitable additional views remain
or the maximum number of steps
is reached.

\paragraph{Render-back feedback:}
After lifting, the resulting Gaussian instance
is rendered into an observation frame
where it is highly visible. Its projected
outline is overlaid on the corresponding
RGB image and returned to the VLM,
together with its geometric properties,
supporting views, and refinement statistics.
This allows the VLM to inspect the lifted
result in image space and, if necessary,
revisit earlier observations, refine the
segmentation, or request further lifting.
The lifted instance is therefore not
automatically accepted as the final answer.

\section{Experiments}
\label{others}



\subsection{\dsname{} Dataset and Evaluation Metrics}
\label{sec:dataset}

\paragraph{Dataset:}
We introduce \dsname{}, a benchmark for part-level
referring segmentation comprising 261 manually verified
queries across 12 ScanNet++ scenes. The queries reference
201 object and 95 part instances annotated on
laser-scanned meshes. They cover five referring roles:
objects, parts as targets, parts as discriminators,
parts as anchors, and multiple targets. In 154 queries
(59\%), parts serve as targets, discriminators, or
anchors. Each query was manually reviewed to ensure
unambiguous target identification and
natural, contextually appropriate wording.
Dataset construction, query statistics, and comparisons
with existing benchmarks are detailed in
Appendix~\ref{app:partref}.

\paragraph{Evaluation metrics:}
We evaluate on 216 held-out test frames, excluded
from all methods' perception and preprocessing.
Predicted Gaussian subsets are rendered within their
complete scenes to obtain occlusion-aware masks,
which are compared against ground-truth masks
ray-traced from the laser-scanned meshes.
For each query, we average the intersection over union (IoU) over test frames
containing at least 300 visible target pixels.
We report mean IoU (mIoU) across queries and the
percentage of queries achieving IoU thresholds
of 0.25 and 0.5 (Acc@0.25/0.5).
Of the 261 queries, 256 are evaluable; the remaining
five have insufficient target visibility.
Further evaluation details are provided in
Appendix~\ref{app:evaluation}.

\subsection{3D Visual Grounding Experiments}
\label{sec:main}

\begin{table}[t]
\vspace{-4mm}
\centering
\small
\caption{
Results on \dsname{}.
Groups: Obj (76), Part (151: part as target,
discriminator or anchor), Multi (29).
Every method uses its original model and candidate source.
GaussDet uses Qwen2.5-VL-7B, OpenGround uses GLM-4.5V
\rev{with its online perception enabled},
and ZeroSplat uses Qwen3-VL-30B\rev{; GPT-5.6 runs with medium reasoning}.
}
\label{tab:main}

\resizebox{\linewidth}{!}{%
\begin{tabular}{l cccc cccc cccc}
\toprule
& \multicolumn{4}{c}{mIoU $\uparrow$}
& \multicolumn{4}{c}{Acc@0.25 $\uparrow$}
& \multicolumn{4}{c}{Acc@0.5 $\uparrow$} \\
\cmidrule(lr){2-5}
\cmidrule(lr){6-9}
\cmidrule(lr){10-13}

Method
& All & Obj & Part & Multi
& All & Obj & Part & Multi
& All & Obj & Part & Multi \\
\midrule
\multicolumn{13}{l}{
\textit{\rev{No VLM: feature retrieval}}
} \\

CLIP retrieval~\citep{clip}
& 0.156 & 0.138 & 0.169 & 0.137
& 0.25 & 0.21 & 0.27 & 0.24
& 0.15 & 0.16 & 0.16 & 0.10 \\

Search3D~\citep{search3d}
& 0.143 & 0.161 & 0.137 & 0.122
& 0.21 & 0.18 & 0.21 & 0.28
& 0.11 & 0.16 & 0.09 & 0.07 \\

\midrule
\multicolumn{13}{l}{
\textit{\rev{Two-stage: a segmenter extracts candidates, then a VLM selects}}
} \\

GaussDet~\citep{gaussdet}
& 0.253 & 0.293 & 0.237 & 0.230
& 0.45 & 0.49 & 0.44 & 0.38
& 0.21 & 0.28 & 0.19 & 0.10 \\

OpenGround~\citep{openground}
& 0.176 & 0.195 & 0.156 & 0.229
& 0.24 & 0.24 & 0.22 & 0.38
& 0.18 & 0.22 & 0.15 & 0.24 \\

ZeroSplat~\citep{zerosplat}
& \rev{0.277} & \rev{0.321} & \rev{0.260} & \rev{0.250}
& \rev{0.51} & \rev{0.58} & \rev{0.50} & \rev{0.41}
& \rev{0.21} & \rev{0.32} & \rev{0.17} & \rev{0.17} \\

\midrule
\multicolumn{13}{l}{
\textit{\rev{Closed loop}: the model decides what to perceive}
} \\

\method{} (Qwen3.5-397B)
& 0.450 & 0.469 & 0.453 & 0.385
& 0.71 & 0.70 & 0.73 & 0.62
& 0.48 & 0.50 & 0.48 & 0.38 \\

\method{} (GPT-5.6)
& \textbf{0.538} & \textbf{0.572} & \textbf{0.519} & \textbf{0.548}
& \textbf{0.84} & \textbf{0.88} & \textbf{0.82} & \textbf{0.86}
& \textbf{0.59} & \textbf{0.62} & \textbf{0.58} & \textbf{0.59} \\

\bottomrule
\end{tabular}%
}
\end{table}

\paragraph{Main results:}
As shown in Table~\ref{tab:main}, \method{} achieves
0.538 mIoU, compared with \rev{0.277} for the strongest
baseline, with substantial gains across all 
query types. These results demonstrate the difficulty
of referring segmentation in large indoor scenes,
where identifying the correct target requires
scene exploration and disambiguation. The consistent
improvements support the effectiveness of
query-driven, closed-loop perception.
\rev{Results for each part role are reported in
Table~\ref{tab:roles} (Appendix~\ref{app:roles}).}

Since ZeroSplat and GaussDet had not released
their implementations at the time of our experiments,
we reproduced both methods from their papers.
Their reproduced results on Ref-LERF are reported
in Appendix~\ref{app:baseline_repro}.

\newcommand{\nr}{\textcolor{gray}{--}}   

\begin{table*}[t]
\vspace{-2mm}
\centering
\caption{
Results on external benchmarks (mIoU, \%).
LERF uses 2D test-view masks, while ScanNet
uses 3D vertex labels.
See Appendix~\ref{app:external} for protocols.
\rev{\method{} uses GPT-5.6 with medium reasoning; the other methods use their default models.}
}
\label{tab:ext_benchmarks}

\vspace{3pt}
\scriptsize
\renewcommand{\arraystretch}{1.05}

\begin{minipage}[t]{0.19\textwidth}
\vspace{0pt}
\centering
\textbf{Ref-LERF}
\par\vspace{3pt}
\begin{tabular*}{\linewidth}
{@{}l@{\extracolsep{\fill}}r@{}}
\toprule
ReferSplat  & 29.2 \\
GaussDet    & 45.9 \\
ZeroSplat   & 32.7 \\
CausalSplat & 36.1 \\
\midrule
\method{}   & \textbf{61.1} \\
\bottomrule
\end{tabular*}
\end{minipage}%
\hfill
\begin{minipage}[t]{0.19\textwidth}
\vspace{0pt}
\centering
\textbf{Causal-LERF}
\par\vspace{3pt}
\begin{tabular*}{\linewidth}
{@{}l@{\extracolsep{\fill}}r@{}}
\toprule
ReferSplat       & 10.0 \\
LUDVIG           & 23.6 \\
InstanceGaussian & 6.8 \\
CausalSplat      & 47.0 \\
\midrule
\method{}        & \textbf{65.7} \\
\bottomrule
\end{tabular*}
\end{minipage}%
\hfill
\begin{minipage}[t]{0.19\textwidth}
\vspace{0pt}
\centering
\textbf{GR-LERF}
\par\vspace{3pt}
\begin{tabular*}{\linewidth}
{@{}l@{\extracolsep{\fill}}r@{}}
\toprule
ReferSplat       & 8.8 \\
LUDVIG           & 23.8 \\
InstanceGaussian & 9.9 \\
ZeroSplat        & 50.8 \\
\midrule
\method{}        & \textbf{63.3} \\
\bottomrule
\end{tabular*}
\end{minipage}%
\hfill
\begin{minipage}[t]{0.19\textwidth}
\vspace{0pt}
\centering
\textbf{Causal-ScanNet}
\par\vspace{3pt}
\begin{tabular*}{\linewidth}
{@{}l@{\extracolsep{\fill}}r@{}}
\toprule
OpenGaussian     & 2.9 \\
LUDVIG           & 5.1 \\
InstanceGaussian & 1.4 \\
CausalSplat      & 14.9 \\
\midrule
\method{}        & \textbf{23.6} \\
\bottomrule
\end{tabular*}
\end{minipage}%
\hfill
\begin{minipage}[t]{0.19\textwidth}
\vspace{0pt}
\centering
\textbf{GR-ScanNet}
\par\vspace{3pt}
\begin{tabular*}{\linewidth}
{@{}l@{\extracolsep{\fill}}r@{}}
\toprule
OpenGaussian     & 19.8 \\
LUDVIG           & 15.1 \\
InstanceGaussian & 24.5 \\
ZeroSplat        & \textbf{41.2} \\
\midrule
\method{}        & 24.0 \\
\bottomrule
\end{tabular*}
\end{minipage}

\end{table*}

\paragraph{Evaluation on other datasets:}
We evaluate \method{} on five existing benchmarks
following their protocols (Appendix~\ref{app:external}).
As shown in Table~\ref{tab:ext_benchmarks},
\method{} outperforms published baselines on
four of the five benchmarks.
The exception is GR-ScanNet, where \method{}
achieves 24.0 mIoU versus ZeroSplat's 41.2.
A likely factor is the mismatch in evaluation
granularity: GR-ScanNet uses category-level
ground truth, which can include multiple
instances of the same category, whereas
\method{} identifies the specific instances
referred to by the query. Consequently,
instance-level predictions may be penalized
for not covering other same-category regions.

\subsection{Effect of Model Capability}
\label{sec:role}

Table~\ref{tab:role} compares \rev{five} VLM backbones
across different perception strategies.
Changing the backbone has \rev{no consistent} effect on
ZeroSplat (\rev{0.222--0.325} mIoU) or OpenGround
(0.161--0.180). In contrast, \method{}
improves from 0.128 to \rev{0.538} mIoU.

Ground-truth(GT) interventions further expose
the limitations of candidate-based grounding.
Supplying ground-truth candidate masks to OpenGround improves to \rev{at most 0.441}, while making oracle selection over an augmented candidate pool collected across four runs achieves 0.435.
ZeroSplat reaches 0.293 when its text
and box prompts are provided from ground truth, \rev{and up to 0.437 when its segmentation masks
are replaced by ground truth} (Appendix~\ref{app:role}). With Qwen3.5-397B \method{} \rev{(0.450)} outperforms both baselines, even when they use ground truth information \rev{(at most 0.441)}.

These results suggest that stronger VLMs
are more effectively utilized when they
control perception rather than only selecting
from generated candidates. The poor
performance of \method{} with the 30B model
also highlights the importance of effective
tool use.

\begin{table}[t]
\vspace{-4mm}
\centering
\footnotesize
\setlength{\tabcolsep}{3.5pt}
\renewcommand{\arraystretch}{1.08}

\caption{
Effect of VLM choice and perception strategy
on \dsname{} (mIoU).
$\Delta$ denotes the change from Qwen3-VL-30B
to \rev{GPT-5.6, which runs with medium reasoning}.
GT masks and GT table replace the respective
methods' segmentation masks and candidate table.
The final column reports GT-guided prompting
for ZeroSplat and oracle selection for OpenGround.
See Appendix~\ref{app:role}.
}
\label{tab:role}

\begin{tabular}{@{}l ccccc rc@{}}
\toprule
& Qwen3-VL & Qwen3.5 & Qwen3.5 & \rev{Qwen3.8} & & & GT \\
Method
& 30B
& 35B
& 397B
& \rev{Flash}
& \rev{GPT-5.6}
& $\Delta$
& intervention \\
\midrule

ZeroSplat
    & \rev{0.277} & \rev{0.269} & \rev{0.288} & \rev{0.222} & \rev{0.325}
    & \rev{$+0.048$} & 0.293 \\

\quad + GT masks
    & \rev{0.390} & \rev{0.402} & \rev{0.415} & \rev{0.374} & \rev{0.437}
    & \rev{$+0.047$} & -- \\

\addlinespace[3pt]

OpenGround
    & 0.161 & 0.173 & 0.180 & \rev{0.174} & \rev{0.162}
    & \rev{$+0.001$} & 0.435 \\

\quad + GT table
    & 0.349 & 0.360 & 0.424 & \rev{0.417} & \rev{0.441}
    & \rev{$+0.092$} & -- \\

\midrule

\method{}
    & 0.128 & 0.353 & 0.450 & \rev{0.492} & \rev{\textbf{0.538}}
    & \rev{$\mathbf{+0.410}$} & -- \\

\bottomrule
\end{tabular}
\vspace{-10pt}
\end{table}

\begin{table}[t]
\centering
\footnotesize
\setlength{\tabcolsep}{5pt}
\renewcommand{\arraystretch}{1.05}

\caption{
Ablation and oracle results on \dsname{}
using Qwen3.5-397B.
Wrong counts queries with IoU below 0.1.
\rev{One-pass segments up to 200 frames with one phrase and lets the VLM
select once; No 3D in the loop refines masks in 2D only, and the answered masks are
lifted once at the end with the same Lifter as in the full system.}
Details in App.~\ref{app:oracle}.
}
\label{tab:ablation}

\begin{tabular}{@{}lccccc@{}}
\toprule
Setting & Obj & Part & Multi & All & Wrong \\
\midrule

\multicolumn{6}{@{}l}{
\textit{Interaction and feedback ablations}
} \\

One look, one cut
& 0.224 & 0.199 & 0.215 & 0.209 & 154 \\

\rev{One-pass}
& \rev{0.444} & \rev{0.398} & \rev{0.404} & \rev{0.412} & \rev{79} \\

No visual exploration
& 0.344 & 0.301 & 0.348 & 0.319 & 114 \\

No segmentation feedback
& 0.433 & 0.403 & 0.407 & 0.412 & 68 \\

No lifting feedback
& 0.471 & 0.429 & 0.420 & 0.440 & 62 \\

\rev{No 3D in the loop}
& \rev{0.389} & \rev{0.396} & \rev{0.401} & \rev{0.395} & \rev{86} \\

\midrule
Full \method{}
& 0.469 & 0.453 & 0.385 & \textbf{0.450} & 55 \\
\midrule

\multicolumn{6}{@{}l}{
\textit{Progressive ground-truth interventions}
} \\

GT boxes, same frames
& 0.591 & 0.558 & 0.585 & 0.571 & 12 \\

GT boxes, GT frames
& 0.595 & 0.563 & 0.616 & 0.578 & 13 \\

GT masks, same $K$ frames
& 0.691 & 0.682 & 0.722 & 0.689 & 6 \\

\bottomrule
\end{tabular}
\vspace{-10pt}
\end{table}

\subsection{Ablations and Oracle Experiments}
\label{sec:ablation}

Table~\ref{tab:ablation} examines the role of
closed-loop interaction and individual
perceptual feedback channels.
Restricting the system to a single round observation
and segmentation decision reduces mIoU
from 0.450 to 0.209.
\rev{A one-pass pipeline that follows the two-stage process with our own tools,
segmenting up to 200 frames with a phrase given by the VLM and letting the VLM
select once among the lifted candidates, reaches 0.412 with this backbone and
0.458 with GPT-5.6, against 0.450 and 0.538 for the full system.}
Among the three feedback ablations,
removing visual exploration causes the
largest drop, to 0.319 mIoU.
Suppressing segmentation and lifting feedback
yields 0.412 and 0.440 mIoU, respectively.
\rev{Lifting feedback is only part of what 3D contributes: removing 3D from the loop
altogether, so that the VLM refines masks in 2D only and they are lifted once at the
end, yields 0.395.}
These results highlight the importance of
early visual observations and suggest that
the remaining feedback channels can partially
compensate for missing information.
We analyze the resulting tool-use behavior
in Appendix~\ref{app:oracle}.

Oracle experiments reveal further opportunities
for improvement. Ground-truth boxes increase
mIoU to 0.571, while ground-truth frame selection
provides only a marginal additional gain (0.578).
Replacing segmentation masks with ground truth
further improves performance to 0.689,
highlighting the importance of accurate
query-directed segmentation.

\FloatBarrier
\section{Conclusion}

We introduced \method{}, a query-driven,
closed-loop approach to part-level referring
segmentation in Gaussian scenes. By placing
perception under VLM control, our method
enables the model to select observations,
construct query-specific instances, and
refine its predictions through perceptual
feedback without precomputed semantic candidates.
We also introduced \dsname{}, a benchmark
covering diverse part-level referring expressions
with mesh-derived ground truth.
\method{} substantially outperforms existing
methods on our benchmark and achieves higher
scores on four external referring benchmarks.
Our experiments further demonstrate the
benefits of iterative interaction and show
that feedback from different perceptual
stages plays complementary roles.
These findings highlight the potential
of closed-loop perception for leveraging
increasingly capable VLMs in fine-grained
3D scene understanding.

\begin{ack}
This work was supported by the King Abdullah University of
Science and Technology (KAUST) research fund under grants
BAS/1/1680-01-01 and BAS/1/1695-01-01.
\end{ack}

\bibliographystyle{plainnat}
\bibliography{references}

\appendix

\section{Closed-Loop Grounding Implementation}
\label{app:loop}

\paragraph{Tool interface:}
Table~\ref{tab:tools} lists the tools available
to the VLM, their arguments, and the feedback
returned after execution.

\begin{table}[H]
\centering
\footnotesize
\setlength{\tabcolsep}{3pt}
\renewcommand{\arraystretch}{1.08}
\caption{
Tool interfaces of \method{}, organized
by tool set. Optional arguments are
shown in brackets.
}
\label{tab:tools}

\begin{tabular}{@{}l p{0.25\linewidth} p{0.53\linewidth}@{}}
\toprule
Action & Arguments & Feedback \\
\midrule

\multicolumn{3}{@{}l}{\textbf{Navigator}} \\[2pt]

\texttt{explore}
& $n$
& Unseen frames selected by additional coverage \\

\texttt{nearby}
& $f$
& Neighboring frames \\

& $f$, box
& Full-resolution crop and frames observing
the selected region \\

& $s$ or $e$
& Alternative views of a group or instance \\

\addlinespace[3pt]
\midrule
\multicolumn{3}{@{}l}{\textbf{Segmenter}} \\[2pt]

\texttt{segment}
& $f$, box, word
[, frames, parent]
& Segmentation masks, overlays, cutouts,
box agreement, and alternative matches \\

\texttt{same}
& groups
& Merged object ID and pairwise 3D overlap \\

\addlinespace[3pt]
\midrule
\multicolumn{3}{@{}l}{\textbf{Lifter}} \\[2pt]

\texttt{lift}
& $s$ or $o$
[, thorough]
& Lifted instance ID, rendered inspection
view, and single-view flag \\

\addlinespace[3pt]
\midrule
\multicolumn{3}{@{}l}{\textbf{VLM termination actions}} \\[2pt]

\texttt{answer}
& ids, evidence
& Final Gaussian subset \\

\texttt{report}
& status, detail
[, candidates]
& Termination report \\

\bottomrule
\end{tabular}
\end{table}

\paragraph{Grounding algorithm:}
Algorithm~\ref{alg:loop} describes how the VLM
selects these tools, incorporates feedback,
and terminates with an answer or failure report.
\begin{algorithm}[t]
\caption{Closed-Loop Grounding in \method{}}
\label{alg:loop}
\begin{algorithmic}[1]
\Require Query $Q$, scene $\mathcal{G}$,
    frames $\mathcal{F}$, VLM $\pi$,
    tools $\Phi$, budget $B$
\Ensure Gaussian subset $\mathcal{S}$
    or a failure report

\State $E_0,\mathcal{H}_0,\tau_0
    \gets\emptyset,\emptyset,()$

\For{$t=0,1,\ldots$}
    \State $\ell_t\gets
        \mathrm{status}(t,E_t,\mathcal{H}_t,B)$
    \If{budget exhausted}
        \State Restrict actions to
            \texttt{answer} and \texttt{report}
    \EndIf

    \State $a_t\gets\pi(Q,\tau_t,\ell_t)$
    \State $E_{t+1},\mathcal{H}_{t+1}
        \gets E_t,\mathcal{H}_t$

    \If{$a_t=\texttt{answer}(I,\varepsilon)$}
        \If{$\mathrm{legal}(I,\varepsilon,E_t)$}
            \State \Return
                $\bigcup_{e\in I}\mathcal{G}_e$
        \EndIf
        \State $r_t\gets
            \mathrm{constraint\_feedback}$
    \ElsIf{$a_t=\texttt{report}(\cdot)$}
        \State \Return failure report
    \ElsIf{$a_t$ is a permitted tool call}
        \State $(r_t,E_{t+1},\mathcal{H}_{t+1})
            \gets$
        \Statex \hspace{\algorithmicindent}
            $\Phi(a_t;E_t,\mathcal{H}_t,
            \mathcal{G},\mathcal{F})$
    \Else
        \State $r_t\gets
            \mathrm{action\_feedback}$
    \EndIf

    \State $\tau_{t+1}\gets
        \tau_t\mathbin{\Vert}(a_t,r_t)$

    \If{two forced termination attempts fail}
        \State \Return \texttt{not\_found}
    \EndIf
\EndFor
\end{algorithmic}
\end{algorithm}

\FloatBarrier
\subsection{\rev{Running Time and Cost}}
\label{app:cost}

\rev{Most compared methods do not report their time per query, so we measured them under one
setting: one RTX PRO 5000 Blackwell GPU and GPT-5.6 with medium reasoning.
Table~\ref{tab:cost} reports the averages per query.}

\begin{table}[H]
\centering
\small
\caption{\rev{Average cost per query with GPT-5.6 ($^{*}$ estimated).}}
\label{tab:cost}
\begin{tabular}{@{}lcccccc@{}}
\toprule
Method & mIoU & VLM calls & Tokens in & Tokens out & SAM3 frames & Time \\
\midrule
OpenGround & 0.162 & 6.9 & 23k & 2.7k & 0.1 & 89 s \\
ZeroSplat & 0.325 & 30.7 & 53k & 7.2k & 247 & 208 s \\
TAB & -- & $>$450$^{*}$ & $>$1M$^{*}$ & -- & $>$300$^{*}$ & $>$30 min$^{*}$ \\
\method{} & \textbf{0.538} & 9.3 & 74k & 1.3k & 26 & 64 s \\
\bottomrule
\end{tabular}
\end{table}

\paragraph{\rev{Input tokens of \method{}:}}
\rev{Every step sends the interaction history, so most of the input hits the prompt cache:
of the 74k input tokens per query, only about 12k are billed at full price.}

\paragraph{\rev{OpenGround:}}
\rev{Its online extension is enabled but rarely triggered: following the official prompt,
GPT-5.6 maps a part to its parent's label, so no segmentation is requested.}

\paragraph{\rev{TAB:}}
\rev{TAB~\citep{tab} could not be run on the full benchmark; its row is an estimate.
It queries the VLM on every frame in which the target category is segmented, and a part
such as a handle is visible in almost every frame.}

\section{PartRef-ScanNet++: Dataset and Evaluation Details}
\label{app:partref}

\subsection{Benchmark Comparison and Statistics}
\label{app:dataset_compare}

\paragraph{Comparison with existing benchmarks:}
Table~\ref{tab:dataset_compare} compares \dsname{}
with existing referring benchmarks. Previous
Gaussian-based benchmarks primarily focus on
object-level grounding. OpenTarget~\citep{openground}
extends 3D grounding to articulated parts using
3D bounding-box annotations. In contrast, \dsname{}
covers both objects and parts, including expressions
in which parts serve as targets, discriminators,
or relational anchors, and provides mesh-level
segmentation ground truth.

\begin{table}[htbp]
\centering
\small
\caption{Comparison with existing referring benchmarks.}
\label{tab:dataset_compare}
\resizebox{\linewidth}{!}{
\begin{tabular}{lrrcccl}
\toprule
& & & \multicolumn{3}{c}{Targets} & \\
\cmidrule(lr){4-6}
Benchmark & Scenes & Queries & Object & Part & Multi
& Ground truth \\
\midrule
Ref-LERF~\citep{refersplat}
    & 4 & 59 & \checkmark & -- & -- & 2D masks \\
GR-LERF~\citep{zerosplat}
    & 4 & 111 & \checkmark & -- & \checkmark & 2D masks \\
GR-ScanNet~\citep{zerosplat}
    & 10 & 83 & \checkmark & -- & \checkmark & Point labels \\
Causal-LERF~\citep{causalsplat}
    & 4 & 158 & \checkmark & -- & -- & 2D masks \\
Causal-ScanNet~\citep{causalsplat}
    & 10 & 73 & \checkmark & -- & -- & Point labels \\
OpenTarget~\citep{openground}
    & 236 & 7,724 & -- & \checkmark & -- & 3D boxes \\
\midrule
\textbf{\dsname{}}
    & \textbf{12} & \textbf{261}
    & \checkmark & \checkmark & \checkmark
    & 3D mesh masks \\
\bottomrule
\end{tabular}
}
\end{table}

\paragraph{Comparison with OpenTarget:}
OpenTarget contains 7,724 descriptions across
236 ScanNet++ scenes, with part annotations from
Articulate3D. Its descriptions are generated by
a VLM and subsequently reviewed by human annotators.
While OpenTarget focuses on articulated parts and
uses 3D bounding boxes for evaluation, \dsname{}
additionally includes whole objects, non-articulated
parts, and multi-target expressions, evaluated
using mesh-derived segmentation masks.

OpenTarget contains substantially longer
descriptions, averaging approximately 31 words,
compared with a median of nine words in \dsname{}.
Our annotation process emphasizes natural,
contextually appropriate expressions, with
manual review to ensure that each query
unambiguously identifies its intended target.

\paragraph{Evaluation on OpenTarget:}
To further evaluate generalization, we compare
\method{} with OpenGround on OpenTarget.
We sample 300 of the 651 OpenTarget queries
from the 12 scenes shared with \dsname{},
using only training frames and OpenGround's
official evaluation protocol.

Both methods use Qwen3.5-397B. For \method{},
we convert the predicted Gaussian subset
into a 3D bounding box using the primitives
accounting for 99\% of its mass.
Two queries blocked by API content filtering
are excluded from both methods.

As shown in Table~\ref{tab:opentarget},
\method{} achieves higher box accuracy
under both IoU thresholds.

\begin{table}[htbp]
\centering
\small
\caption{Evaluation on OpenTarget using 300 sampled
queries from the 12 shared scenes.
Two blocked queries are excluded from evaluation.
Both methods use Qwen3.5-397B.}
\label{tab:opentarget}
\begin{tabular}{@{}lcc@{}}
\toprule
Method & Acc@0.25 & Acc@0.5 \\
\midrule
OpenGround & 10.0 & 6.3 \\
\method{} & \textbf{33.9} & \textbf{22.5} \\
\bottomrule
\end{tabular}
\end{table}

\begin{figure}[htbp]
    \centering
    \includegraphics[width=0.65\linewidth]
    {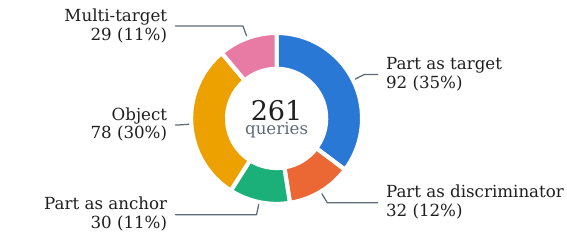}
    \caption{Distribution of the 261 queries
    in \dsname{} across five referring roles.}
    \label{fig:roles}
\end{figure}

\paragraph{Dataset statistics:}
\dsname{} contains 261 queries across 12 ScanNet++
scenes, covering five referring roles:
78 object targets, 92 part targets,
32 part discriminators, 30 part anchors,
and 29 multi-target queries.
As shown in Fig.~\ref{fig:roles},
154 queries (59\%) involve parts as
targets, discriminators, or anchors.

\begin{figure}[htbp]
\centering
\footnotesize
\setlength{\tabcolsep}{0pt}
\begin{tabular}{@{}c@{\hspace{4pt}}c@{\hspace{6pt}}c@{}}
\rotatebox[origin=c]{90}{\textbf{Object}} &
\raisebox{-0.5\height}{\includegraphics[width=0.455\linewidth]{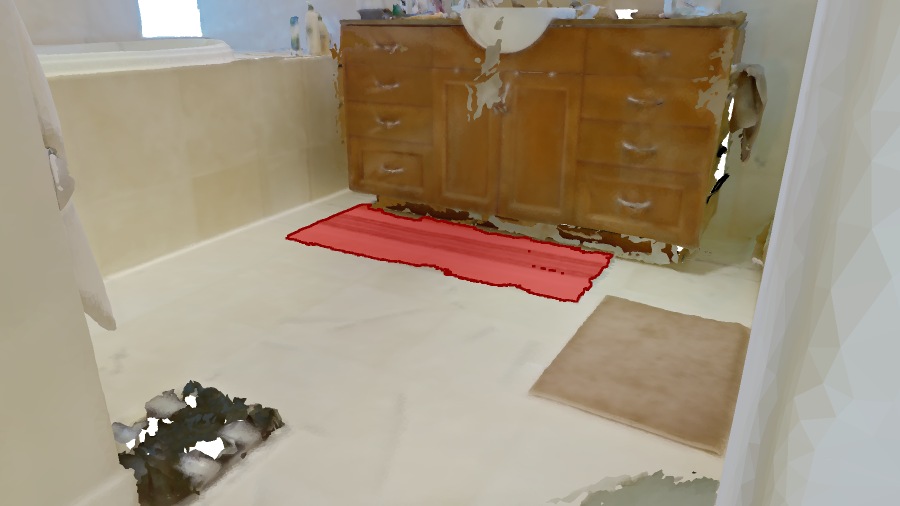}} &
\raisebox{-0.5\height}{\includegraphics[width=0.455\linewidth]{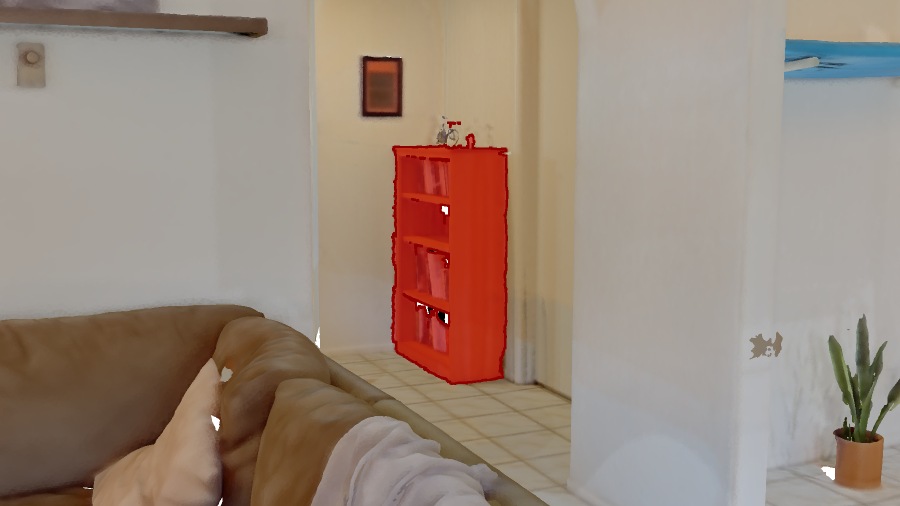}} \\
\noalign{\vskip 2pt}
& {\scriptsize ``the rug lying up against the foot of the vanity''} & {\scriptsize ``the wooden bookshelf standing in the tiled hallway''} \\
\noalign{\vskip 5pt}
\rotatebox[origin=c]{90}{\textbf{Part as target}} &
\raisebox{-0.5\height}{\includegraphics[width=0.455\linewidth]{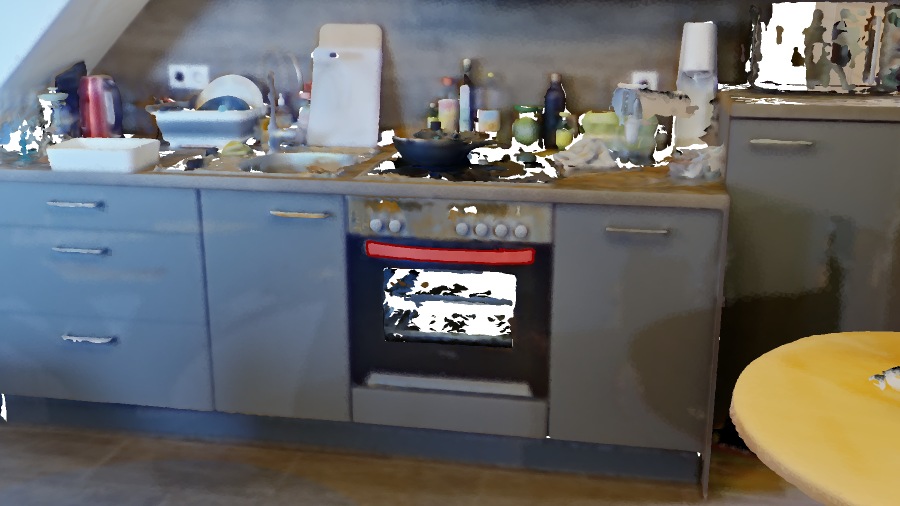}} &
\raisebox{-0.5\height}{\includegraphics[width=0.455\linewidth]{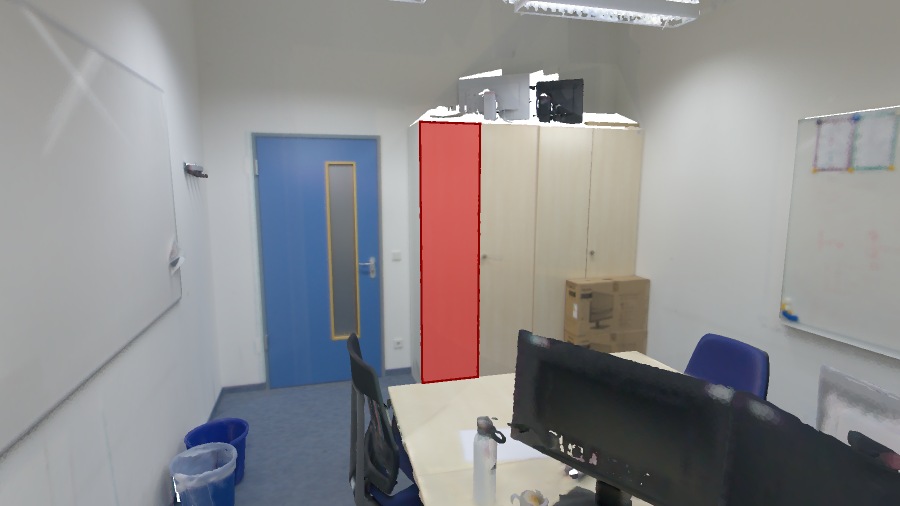}} \\
\noalign{\vskip 2pt}
& {\scriptsize ``the handle of the oven below the cooktop''} & {\scriptsize ``the cabinet door nearest to the blue entrance door''} \\
\noalign{\vskip 5pt}
\rotatebox[origin=c]{90}{\textbf{Part as discriminator}} &
\raisebox{-0.5\height}{\includegraphics[width=0.455\linewidth]{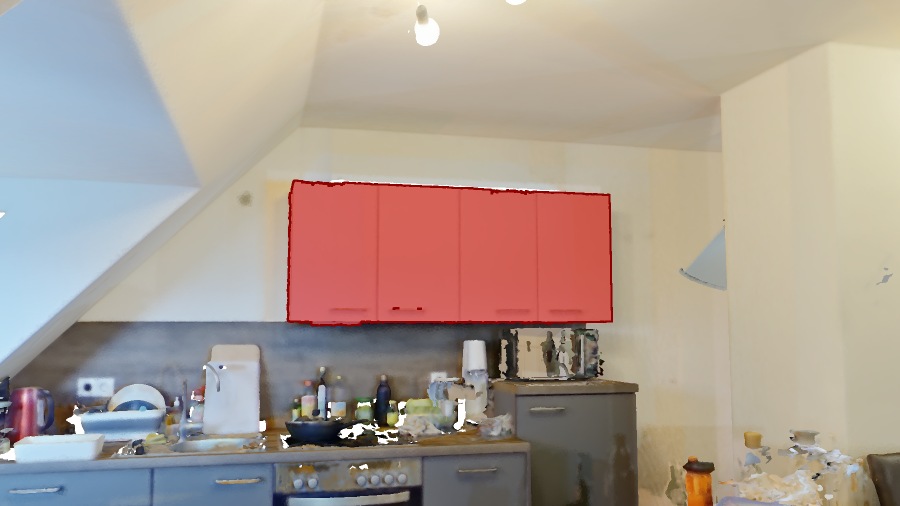}} &
\raisebox{-0.5\height}{\includegraphics[width=0.455\linewidth]{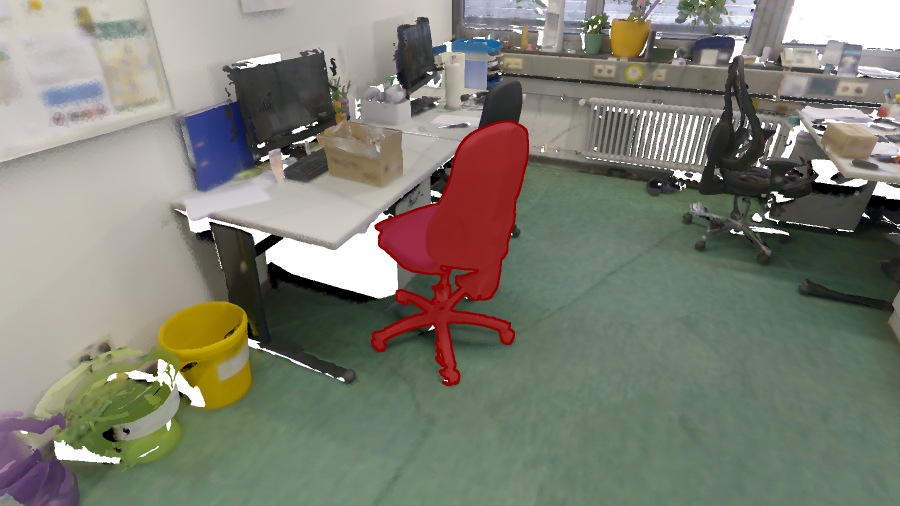}} \\
\noalign{\vskip 2pt}
& {\scriptsize ``the cabinet with the four doors in a row''} & {\scriptsize ``the office chair with the blue seat cushion''} \\
\noalign{\vskip 5pt}
\rotatebox[origin=c]{90}{\textbf{Part as anchor}} &
\raisebox{-0.5\height}{\includegraphics[width=0.455\linewidth]{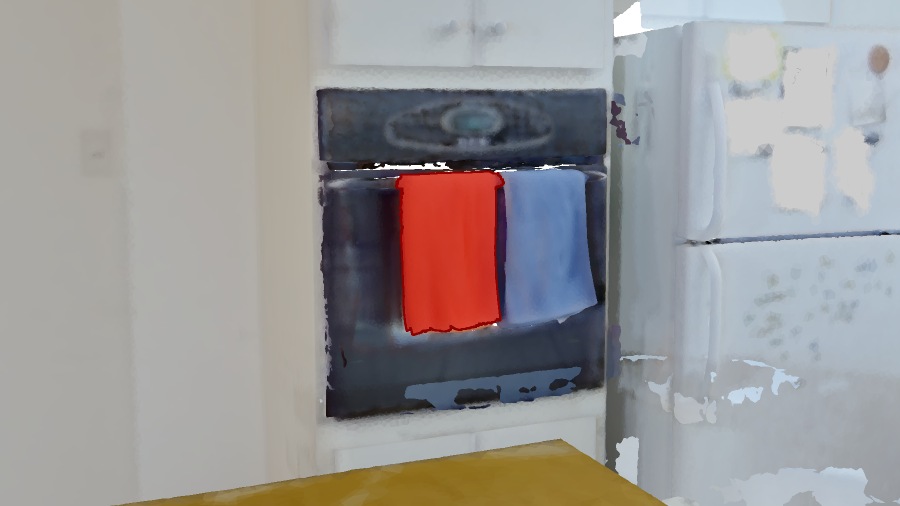}} &
\raisebox{-0.5\height}{\includegraphics[width=0.455\linewidth]{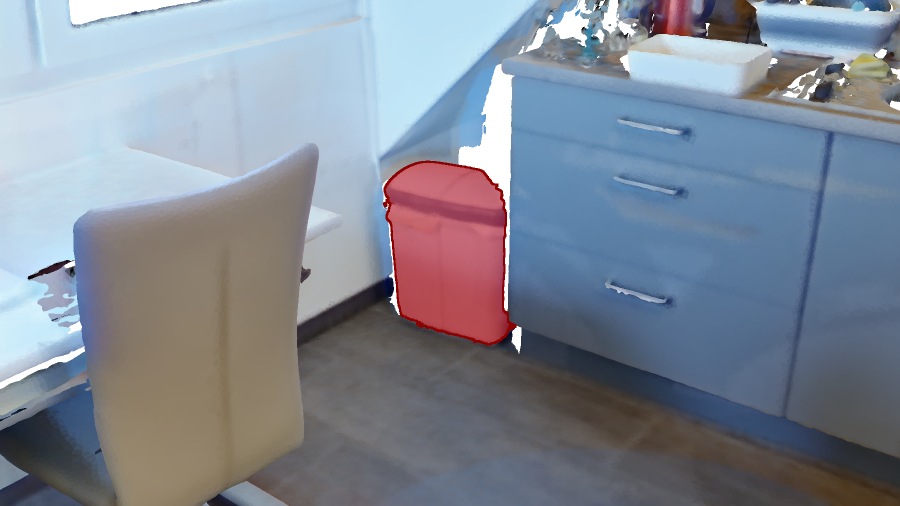}} \\
\noalign{\vskip 2pt}
& {\scriptsize ``the orange towel hanging on the oven handle''} & {\scriptsize ``the bin standing beside the drawers in the kitchen''} \\
\noalign{\vskip 5pt}
\rotatebox[origin=c]{90}{\textbf{Multiple targets}} &
\raisebox{-0.5\height}{\includegraphics[width=0.455\linewidth]{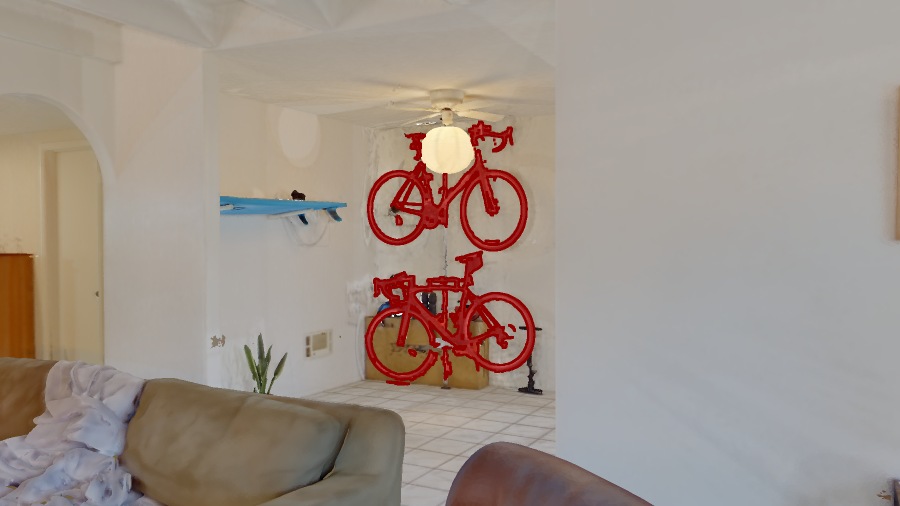}} &
\raisebox{-0.5\height}{\includegraphics[width=0.455\linewidth]{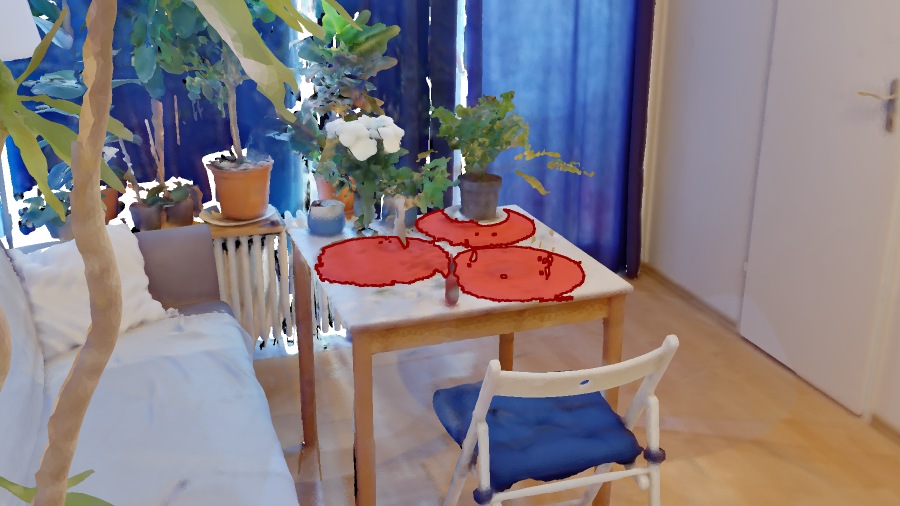}} \\
\noalign{\vskip 2pt}
& {\scriptsize ``the bicycles mounted on the wall''} & {\scriptsize ``the round woven placemats lying on the dining table''} \\
\end{tabular}
\caption{\rev{Example queries of \dsname{} for each referring role. The ground-truth target of each query
is shown in red on the laser-scanned mesh.}}
\label{fig:query_examples}
\end{figure}

\rev{Fig.~\ref{fig:query_examples} shows two example queries of each role with their ground-truth targets.}

\subsection{Dataset Construction}
\label{app:dataset_construction}

\paragraph{Scenes and annotations:}
We select 12 ScanNet++ scenes, each providing
laser-scanned meshes and posed DSLR images.
Following the official splits, we reconstruct
one 2DGS scene using the training frames of
each scene. The 216 held-out test frames are
reserved exclusively for evaluation.

Object annotations are obtained from ScanNet++,
yielding 1,616 valid instances after filtering
non-object labels. Part annotations are derived
from 723 Articulate3D\citep{articulate3d} instances and supplemented
with 24 manually verified parts. All annotations
are represented as vertex subsets of the same
laser-scanned meshes. The final queries reference
201 object and 95 part instances.

\paragraph{Query construction and verification:}
Initial query candidates are generated with AI assistance from mesh-derived geometric relations and subsequently reviewed and revised by human annotators.
Scene-level relation summaries identify suitable
anchors and spatial relationships that distinguish
target instances. Viewpoint-dependent expressions
are excluded.

Each expression is checked for naturalness,
target uniqueness, and the necessity of its
qualifying conditions. We additionally verify
target visibility and annotation validity.

All 357 initial queries undergo manual review
using a 3D annotation interface displaying
highlighted mesh targets and corresponding
images. Of these, 254 are retained, nine are
revised, and 94 are discarded. After
deduplication, the final benchmark contains
261 queries.

\paragraph{Ground-truth generation:}
For each held-out test frame, we generate
occlusion-aware ground-truth masks by ray tracing
the laser-scanned meshes. A pixel belongs to
the target only when the first ray intersection
lies on its annotated mesh surface.
Privacy-masked regions are excluded from evaluation.

\subsection{Evaluation Protocol}
\label{app:evaluation}

\paragraph{Evaluation setup:}
We evaluate predictions against mesh-derived
ground truth on 216 held-out test frames.
These frames are excluded from all methods'
perception and preprocessing procedures.

The evaluator supports binary Gaussian subsets,
per-Gaussian soft weights, and mesh vertex subsets.
For vertex-based submissions, a triangle is
included when at least two of its vertices
belong to the predicted subset.

\paragraph{Occlusion-aware rendering evaluation:}
For Gaussian-based predictions, the submitted
subset is rendered together with the complete
scene. Each selected Gaussian is assigned a
membership value of one, while all remaining
Gaussians are assigned zero. The membership
values are composited using the original
Gaussian opacities.

A pixel is classified as foreground when
the accumulated target contribution exceeds
0.5 and the total accumulated opacity
exceeds 0.5. For soft-weight predictions,
the submitted weights replace binary
membership values.

The predicted masks are compared against
occlusion-aware ground-truth masks obtained
by ray tracing the laser-scanned meshes.
Only frames containing at least 300
visible ground-truth pixels are evaluated.

For each query, we average the IoU across
all valid test frames:
\begin{equation}
    \operatorname{IoU}_q =
    \frac{1}{|\mathcal{V}_q|}
    \sum_{f\in\mathcal{V}_q}
    \frac{|P_q^f\cap G_q^f|}
         {|P_q^f\cup G_q^f|},
    \label{eq:query_iou}
\end{equation}
where $\mathcal{V}_q$ denotes the valid
test frames for query $q$, and $P_q^f$
and $G_q^f$ are the predicted and
ground-truth masks, respectively.

We report mean IoU (mIoU) across queries,
together with Acc@0.25 and Acc@0.5.
Of the 261 queries, 256 are evaluable;
the remaining five have insufficient
target visibility.

\subsection{\rev{Results by Part Role}}
\label{app:roles}

\rev{Table~\ref{tab:roles} breaks down the Part group of
Table~\ref{tab:main} by the role that the part plays in the query:
target (91 evaluable queries), discriminator (31), and anchor (29).
Each count is one below that of Fig.~\ref{fig:roles}, because one query
of each part role has insufficient target visibility in the test frames
and is not evaluable (Sec.~\ref{sec:dataset}).}

\begin{table}[htbp]
\centering
\small
\caption{\rev{Part-level results on \dsname{} by the role of the part.
Part (151 evaluable queries) pools the three roles: target (91),
discriminator (31), and anchor (29). Methods and models are those of
Table~\ref{tab:main}; with Qwen3.5-397B, two target queries were blocked
by API filtering and are excluded.}}
\label{tab:roles}
\resizebox{\linewidth}{!}{%
\begin{tabular}{l cccc cccc cccc}
\toprule
& \multicolumn{4}{c}{mIoU $\uparrow$}
& \multicolumn{4}{c}{Acc@0.25 $\uparrow$}
& \multicolumn{4}{c}{Acc@0.5 $\uparrow$} \\
\cmidrule(lr){2-5}
\cmidrule(lr){6-9}
\cmidrule(lr){10-13}
Method
& Part & Target & Discr. & Anchor
& Part & Target & Discr. & Anchor
& Part & Target & Discr. & Anchor \\
\midrule
\multicolumn{13}{l}{\textit{No VLM: feature retrieval}} \\
CLIP retrieval~\citep{clip}
    & 0.169 & 0.103 & 0.439 & 0.085
    & 0.27 & 0.18 & 0.68 & 0.10
    & 0.16 & 0.07 & 0.48 & 0.10 \\
Search3D~\citep{search3d}
    & 0.137 & 0.117 & 0.261 & 0.069
    & 0.21 & 0.20 & 0.32 & 0.10
    & 0.09 & 0.05 & 0.26 & 0.03 \\
\midrule
\multicolumn{13}{l}{\textit{Two-stage: a segmenter extracts candidates, then a VLM selects}} \\
GaussDet~\citep{gaussdet}
    & 0.237 & 0.171 & 0.374 & 0.297
    & 0.44 & 0.31 & 0.71 & 0.55
    & 0.19 & 0.11 & 0.29 & 0.34 \\
OpenGround~\citep{openground}
    & 0.156 & 0.098 & 0.326 & 0.155
    & 0.22 & 0.15 & 0.45 & 0.17
    & 0.15 & 0.09 & 0.32 & 0.17 \\
ZeroSplat~\citep{zerosplat}
    & 0.260 & 0.189 & 0.411 & 0.319
    & 0.50 & 0.35 & 0.81 & 0.62
    & 0.17 & 0.10 & 0.32 & 0.24 \\
\midrule
\multicolumn{13}{l}{\textit{Closed loop: the model decides what to perceive}} \\
\method{} (Qwen3.5-397B)
    & 0.453 & 0.405 & 0.539 & 0.510
    & 0.73 & 0.69 & 0.81 & 0.76
    & 0.48 & 0.38 & 0.68 & 0.59 \\
\method{} (GPT-5.6)
    & \textbf{0.519} & \textbf{0.449} & \textbf{0.621} & \textbf{0.631}
    & \textbf{0.82} & \textbf{0.74} & \textbf{0.97} & \textbf{0.93}
    & \textbf{0.58} & \textbf{0.44} & \textbf{0.74} & \textbf{0.83} \\
\bottomrule
\end{tabular}%
}
\end{table}

\section{Reproduction of Unreleased Baselines}
\label{app:baseline_repro}

At the time of our experiments, the implementations
of ZeroSplat and GaussDet were not publicly available.
We therefore reimplemented both methods following
their respective papers. To document the differences
between the published results and our reproductions,
we evaluate both implementations on the four
Ref-LERF scenes.

\begin{table}[htbp]
\centering
\footnotesize
\setlength{\tabcolsep}{4pt}
\caption{Published and reproduced results on
Ref-LERF (mIoU, \%). The reproduced implementations
are used in our comparisons on \dsname{}.}
\label{tab:baseline_repro}

\begin{tabular}{@{}lccccc@{}}
\toprule
Method & Waldo & Ramen & Teatime
       & Figurines & Mean \\
\midrule
ZeroSplat (published)
    & 21.3 & 30.4 & 41.4 & 37.8 & 32.7 \\
ZeroSplat (reproduced)
    & \rev{25.2} & \rev{34.7} & \rev{47.6} & \rev{38.0} & \rev{36.4} \\
\midrule
GaussDet (published)
    & 33.3 & 52.3 & 39.2 & 58.7 & 45.9 \\
GaussDet (reproduced)
    & 36.2 & 46.7 & 40.5 & 37.6 & 40.2 \\
\bottomrule
\end{tabular}
\end{table}

Our reproductions achieve \rev{36.4} and 40.2 mIoU
for ZeroSplat and GaussDet, respectively, compared
with their published results of 32.7 and 45.9.
Performance differences vary across scenes,
with larger discrepancies on Teatime for ZeroSplat
and Figurines for GaussDet. For evaluation on \dsname{}, we use our reproduced
implementations of ZeroSplat and GaussDet. Results
on other benchmarks are taken directly from
their respective papers.

\section{External Benchmark Evaluation Protocols}
\label{app:external}

\paragraph{Experimental setup:}
We evaluate \method{} on five existing benchmarks:
Ref-LERF~\citep{refersplat},
GR-LERF and GR-ScanNet~\citep{zerosplat},
and Causal-LERF and Causal-ScanNet~\citep{causalsplat}.
We use only the training frames of each benchmark,
without modifying our perception or grounding
pipeline, and follow its original evaluation
protocol. All baseline results in
Table~\ref{tab:ext_benchmarks} are taken from
the respective publications.

\paragraph{2D evaluation on LERF:}
Ref-LERF, GR-LERF, and Causal-LERF evaluate
segmentation on the four LERF scenes using
2D mIoU. We render the predicted Gaussian
subsets from the benchmark's evaluation
views and compare the resulting masks
against the provided ground-truth masks,
following each benchmark's scoring protocol.

\paragraph{3D evaluation on ScanNet:}
GR-ScanNet and Causal-ScanNet use point-level
3D mIoU, following the fixed-point evaluation
setup of OpenGaussian.
We initialize one Gaussian per input ScanNet
point and disable densification and position
updates during reconstruction, preserving
one-to-one correspondence between Gaussians
and ground-truth points.
Consequently, predicted Gaussian membership
can be compared directly with the corresponding
ground-truth point labels, without nearest-neighbor
matching or label transfer.

\paragraph{Generalized referring queries:}
GR-LERF and GR-ScanNet additionally include
zero-target and multi-target queries.
Following ZeroSplat, zero-target queries
receive a score of one when the prediction
is empty and zero otherwise.
For the remaining queries, we evaluate
the predicted union of target instances
against the corresponding ground truth.

\section{VLM and Perception Analysis}
\label{app:role}

\paragraph{Experimental setup:}
All methods use only training frames, with fixed
method-specific configurations across VLM backbones.
\method{} uses identical prompts, tools, and
execution budgets across models. ZeroSplat follows
its published settings, while OpenGround uses
\rev{its released code and default configuration, with online perception enabled, and}
a temperature of 0.2. Decoding settings differ
between some ZeroSplat runs, and API filtering
reduces the evaluable queries in a few runs
from 256 to 254--255.

\paragraph{Ground-truth interventions:}
We conduct three distinct interventions.
For ZeroSplat, GT masks replace segmentation
outputs without correcting the VLM's instance
selection. \rev{Performance rises to 0.374--0.437 mIoU across the five backbones.} Separately, GT-guided prompting
supplies the correct category and ground-truth
bounding boxes while retaining ZeroSplat's
segmentation and lifting pipeline, reaching
0.293 mIoU.

For OpenGround, GT table replaces its candidates
with official ScanNet++ instances, reaching
\rev{up to 0.441} mIoU. However, this table has limited
part coverage. Oracle selection instead chooses
the highest-overlap candidate from OpenGround's
static table augmented with candidates recovered
across four runs, reaching 0.435 mIoU.
This oracle is restricted to one candidate
per query, including multi-target queries.

\paragraph{Tool-use behavior:}
The 30B model struggles to use \method{}'s
feedback effectively: it reaches the execution
budget without a valid terminal action on
65 queries, compared with \rev{10} for 397B.
After unsuccessful segmentation, the respective
re-segmentation rates are 52\% and 69\%.
These observations help explain the stronger
backbone's larger gains under closed-loop
perception.

\section{Ablation and Oracle Details}
\label{app:oracle}

\paragraph{Interaction and feedback ablations:}
All experiments use Qwen3.5-397B with the
same tool interfaces and execution budget,
except for the specified restrictions.
In the one-look, one-cut setting, the VLM
receives four overview images and makes
a single frame selection and segmentation
decision without further interaction.
Automatic multi-view lifting remains enabled.
\rev{In the one-pass setting, the VLM gives a phrase from the query, SAM3 segments
up to 200 frames with it, the masks are grouped and lifted in 3D with our tools,
and the VLM selects once among the candidates; two queries blocked by API
filtering are counted as failures. With GPT-5.6
as the selecting model and the same phrases, it reaches 0.458 against 0.538.}

For the feedback ablations, we suppress
one feedback channel at a time while
preserving the other two.
\textit{No visual exploration} prevents
the Navigator from returning images and
disables box-based local inspection,
while retaining frame identifiers and
other necessary information.
\textit{No segmentation feedback} hides
segmentation visualizations and quality
diagnostics, returning only the identifiers
and status needed for subsequent operations.
\textit{No lifting feedback} similarly
suppresses rendered inspection views
and lifting-quality feedback while
preserving the underlying lifting operations.
The VLM remains free to invoke all
permitted tools and revise its decisions.
\rev{\textit{No 3D in the loop} goes further and removes every path from masks to 3D
and back: no lifting, no grouping of masks across views in 3D, and no search for
further views from a mask or a box. The VLM can still select views, zoom in,
segment with a phrase and a box, inspect the masks, segment again, and track a
mask across frames. It answers with the mask groups it has built. Each of them is then
lifted once with the iterative lifting that the full system uses for its final answer
(Sec.~\ref{sec:lifter}), including the refinement rounds over complementary views. The
result is not returned to the VLM, and this step does not count toward the budget.}

\paragraph{Compensation between feedback channels:}
The feedback ablations reveal distinct
patterns of tool use.
Without visual exploration, segmentation
calls increase from 2.5 to 4.6 per query,
including 58 attempts to specify segmentation
boxes in frames not previously visualized
through the Navigator.
The number of queries reported as not found
increases from 15 to 62, indicating that
subsequent feedback cannot fully compensate
for missing early visual observations.

Without segmentation feedback, the VLM
increasingly relies on lifting to inspect
its intermediate results.
Lifting calls increase from 1.8 to 4.9
per query, while the average interaction
length increases from 11.0 to 13.4 steps.
Token consumption also increases from
118k to 131k per query, despite the
reduced segmentation feedback.
This behavior suggests that lifting
feedback can partially substitute for
direct inspection of segmentation masks,
but requires additional interaction.

Removing lifting feedback has a smaller
effect on overall accuracy.
With segmentation feedback still available,
the VLM can continue to inspect and revise
its 2D predictions.
However, part-level mIoU decreases from
0.453 to 0.429, indicating that lifting
feedback retains value for some queries.
\rev{This setting only removes the inspection of the rendered instance. Without
any 3D in the loop, mIoU drops to 0.395: the VLM more often gives up, or returns masks
in only one or two frames.}

\begin{table}[t]
\centering
\footnotesize
\setlength{\tabcolsep}{4pt}
\renewcommand{\arraystretch}{1.08}
\caption{
Tool-use behavior under feedback ablations.
Calls and \rev{steps} are averaged per query.
}
\label{tab:feedback_usage}

\begin{tabular}{@{}lcccc@{}}
\toprule
Setting
& Seg./q
& Steps/q
& Not found \\
\midrule

Full \method{}
    & 2.5 & 11.0 & 15 \\

No visual exploration
    & 4.6 & 13.0 & 62 \\

No segmentation feedback
    & 1.9 & 13.4 & 26 \\

No lifting feedback
    & 2.5 & 11.1 & 17 \\

\bottomrule
\end{tabular}
\end{table}

\paragraph{Progressive oracle experiments:}
We progressively replace the model's
decisions with ground truth while retaining
the same lifting pipeline.
\textit{GT boxes} replaces predicted boxes
and parent scopes with mesh-derived
annotations on the model's selected
training frames, while SAM3 performs
segmentation as usual.
\textit{GT frames} additionally selects
the top $K$ training frames by visible
target area, where $K$ matches the number
of distinct segmentation frames in
the original run.
\textit{GT masks} retains exactly these
frames but replaces SAM3 outputs with
mesh-projected ground-truth masks,
including those used during multi-view
refinement.
The comparison between the final two
settings isolates the contribution
of segmentation quality.

\section{\rev{Failure Analysis for Part Targets}}
\label{app:failure}

\begin{figure}[htbp]
\centering
\footnotesize
\setlength{\tabcolsep}{2pt}
\begin{tabular}{@{}ccc@{}}
\includegraphics[width=0.325\linewidth]{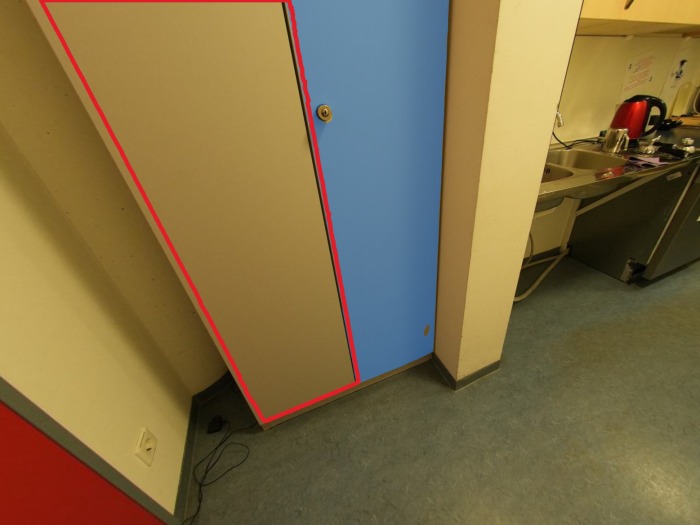} &
\includegraphics[width=0.325\linewidth]{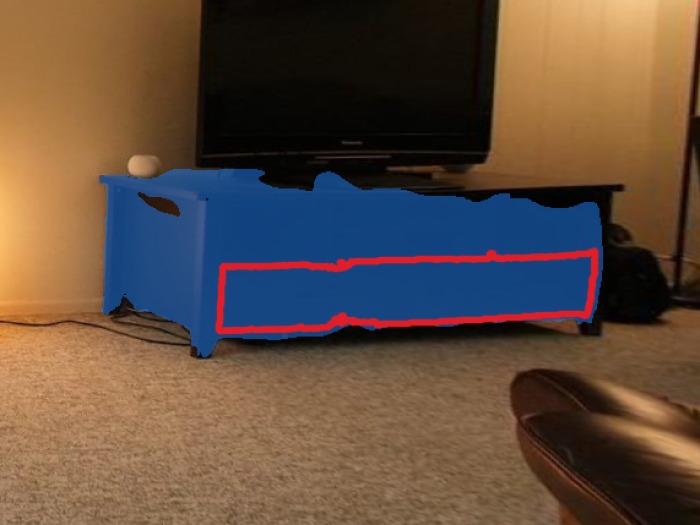} &
\includegraphics[width=0.325\linewidth]{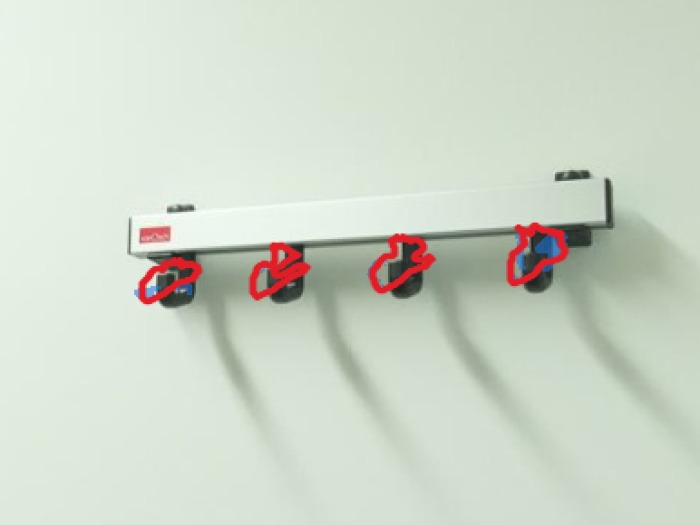} \\[1pt]
\parbox[t]{0.32\linewidth}{\centering ``the door of the tall grey cabinet that hangs nearest the red entrance door'' (IoU~0.03)} &
\parbox[t]{0.32\linewidth}{\centering ``the long drawer at the bottom of the cabinet under the television'' (IoU~0.32)} &
\parbox[t]{0.32\linewidth}{\centering ``the row of black hooks on the wall-mounted coat rack'' (IoU~0.16)} \\
\end{tabular}
\caption{\rev{Failure types on part targets, shown on a test frame with the answer in blue
and the ground truth outlined in red. Left: the answer misses the part and selects the
neighbouring door. Middle: the answer extends beyond the drawer onto the whole cabinet.
Right: the answer covers only a portion of the hooks.}}
\label{fig:failures}
\end{figure}

\rev{We examine the 91 evaluable queries whose target is a part, using the GPT-5.6 run of
Table~\ref{tab:roles} (0.449 mIoU). \method{} reaches an IoU of at least 0.5 on 40 of them.
For the other 51, we compare the rendered answer with the ground truth over the test frames
in terms of pixel precision and recall. Two queries receive no answer.
In 10 queries the answer misses the part (precision and recall both below 0.1), for example by
selecting the neighbouring door of the same cabinet.
In the remaining 39 the answer lies on the target part but is imprecise:
16 answers extend beyond the part (recall above precision), for example onto the whole
cabinet that holds the target drawer, and 23 cover only a portion of it (precision above recall),
10 of which are handles or knobs. Fig.~\ref{fig:failures} shows one example of each type.}

\rev{Table~\ref{tab:failure} groups the queries by the kind of part and by its size.
Handles and knobs are the hardest kind: 61\% of them reach an IoU of 0.25, but only 22\%
reach 0.5. Accuracy also falls with size, from 0.567 mIoU for the largest third of the
parts to 0.353 for the smallest third. The same pattern remains when ground-truth masks are
lifted with our pipeline (GT masks in Table~\ref{tab:ablation}): handles and knobs stay at
0.527 and the smallest third at 0.503. A part of this error therefore comes from lifting small
parts to Gaussians rather than from the decisions of the VLM.}

\begin{table}[htbp]
\centering
\small
\caption{\rev{Part-target queries of \dsname{} by kind and size of the part: \method{} with
GPT-5.6, and the GT masks setting of Table~\ref{tab:ablation} on the same queries.
Size is the mean share of a test image covered by the ground truth; the three size groups
are thirds of the queries.}}
\label{tab:failure}
\begin{tabular}{@{}lc ccc c@{}}
\toprule
& & \multicolumn{3}{c}{\method{}} & GT masks \\
\cmidrule(lr){3-5}\cmidrule(l){6-6}
Group & Queries & mIoU & Acc@0.25 & Acc@0.5 & mIoU \\
\midrule
\multicolumn{6}{@{}l}{\textit{Kind of part}} \\
Handles and knobs & 23 & 0.333 & 0.61 & 0.22 & 0.527 \\
Doors & 15 & 0.536 & 0.80 & 0.67 & 0.738 \\
Drawers & 14 & 0.474 & 0.71 & 0.50 & 0.723 \\
Lids & 13 & 0.560 & 0.92 & 0.54 & 0.714 \\
Seats and backrests & 9 & 0.528 & 0.78 & 0.67 & 0.598 \\
Other & 17 & 0.383 & 0.71 & 0.29 & 0.590 \\
\midrule
\multicolumn{6}{@{}l}{\textit{Size of the part}} \\
Small ($<$0.18\% of the image) & 30 & 0.353 & 0.63 & 0.27 & 0.503 \\
Medium & 30 & 0.424 & 0.77 & 0.37 & 0.611 \\
Large ($>$0.6\% of the image) & 31 & 0.567 & 0.81 & 0.68 & 0.794 \\
\midrule
All & 91 & 0.449 & 0.74 & 0.44 & 0.638 \\
\bottomrule
\end{tabular}
\end{table}

\FloatBarrier

\section{\rev{Controller Prompt and Tool Definitions}}
\label{app:prompt}

\rev{This section lists the instructions and the tool definitions given to the VLM controller
for the tools of Table~\ref{tab:tools}.}

\begin{lstlisting}[style=prompt]
You are locating what a sentence refers to in a 3D indoor scene, and you must answer with instance ids.
You cannot see the scene directly. You have:
- eyes: explore() shows frames not looked at yet in this scene (whole frame, low-res). nearby(frame=) shows other frames overlapping that view. nearby(frame=, box=) zooms: that region at full resolution plus other frames that see the same region. nearby(group=) shows a thing you segmented from other directions.
- a segmenter: segment(frame, box, word, frames=) cuts the thing YOU box on an image you were shown, and follows it into the extra frames you name. It returns the boxed thing as a group marked BOXED, plus look-alikes it found in those frames, with a labeled overlay and white-background cutouts. within= restricts it to inside a parent group or instance (for parts). The segmenter is not a search tool: you decide what gets cut.
- same(groups): declare that several groups are one physical thing -> object id i1. lift(i1) then votes with all their frames.
- lift(group, thorough=): turns a group (s3) or an object (i1) into a 3D instance #id (cheap; thorough=true for the final answer only, it gathers more views by geometry).
- answer(ids, evidence) / report(status, detail, candidates).

Notation: frames f123, segment groups s3, objects i2 (groups you declared to be one thing), instances #12 (lifted). Boxes: box=[x0,y0,x1,y1] on a 0-1000 grid of the image as you saw it (origin top-left); name the frame by the id stamped on that image (in the overview sheet, the id printed on each tile). If the same frame was shown to you more than once, the coordinates refer to the most recent image of it. Group ids only exist after a segment() call returned them. Leave out optional parameters you do not use (never pass empty strings). A [this query] line at the end of each turn lists what you have made so far in this query.

Rules:
1. Re-read the sentence before deciding; never rely on a paraphrase. Every noun in the sentence that constrains the target matters.
2. Nothing is known about the room when a sentence arrives: start by looking. explore() a few whole frames to see the room, then zoom into the region that shows the right area with nearby(frame=, box=) until the thing is clear.
3. Look before you cut. Find the thing with your eyes first: explore to reach frames not looked at yet; nearby(frame=, box=) to zoom on a region until you can draw a tight box around the thing (small things are invisible in whole frames, so zoom before boxing them); nearby(group=) to see more of something already segmented. Then segment it with a tight box. The word is only its label (1-2 plain nouns: 'cup', 'dish rack'); if you see the thing but no word fits, box it anyway. If coverage stops growing and you still have not seen the thing, it is probably absent.
4. Segment and lift only what you need. The target itself: always (the answer must be a lifted instance). Anchors and competing candidates: only when the images leave a relation uncertain. Everything else can stay unsegmented, and an anchor that will not segment (glass, thin, cluttered) is not a blocker when the images already show the relation.
5. Relations between two things visible in the same image may be judged from the image; that is the default. above/below/on mean vertically aligned; higher/lower only compare heights. A single-view lift is fine as an anchor but not as an answer.
6. For a part of a larger thing (a handle, a drawer, a door): zoom on the parent, box the part, and call segment(frame, box, part_word, within=parent) where parent is the parent's group or instance id. Check the overlay.
7. The same thing often comes back as several groups: re-segmented in other frames, or split by the segmenter. Each segment result lists which earlier groups a new group overlaps in 3D; judge with the cutouts and positions, then declare same(groups=[...]) -> i1 (or pass same_as= in segment) and lift the object. A group cut in one frame lifts as SINGLE-VIEW; for anything you will answer with, gather several views. lift(thorough=true) the answer before answering; look at the red outline in the acceptance image. If it is wrong, box it again in a better frame.
8. Before answer(): evidence is one triple per condition in the sentence: {subject:'#5', relation:'next to', object:'#2' or a name, how:'image f12 step 3'}. A condition without evidence -> keep working, or report honestly (ambiguous with the tied candidates, or not_found). Never guess.
9. A sentence can refer to more than one thing: a plural ("the chairs", "the two knife sets"), a whole kind ("objects used to store office supplies", "every cabinet"), or several named things. Then the answer is ALL the instances that fit, each found, segmented and lifted on its own, and answer(ids=[...]) lists every one of them; stop only when you have checked that no further instance of that kind is visible in the frames. Answering one of several is wrong.
10. Budget: at most 16 tool calls. Be frugal: one explore is usually enough to find a large anchor; do not re-segment the same thing with other words hoping for a different answer.
\end{lstlisting}

\begin{lstlisting}[style=prompt]
explore(n)
  Show n frames not looked at yet in this scene, chosen to add the most new scene area. Returns the frames (whole, low-res) and the session coverage.
  - n (integer, optional): how many frames, 1-4 (default 4)

nearby(frame, box, group, n)
  Look closer. Three uses, give exactly one: frame='f123' shows other frames overlapping that view (what else is around); frame='f123' plus box=[x0,y0,x1,y1] zooms on that region of the frame at full resolution and shows other frames that see the same region from other directions; group='s3' / 'i1' / '#12' shows crops of that segmented thing or lifted instance from other directions, outlined.
  - frame (string, optional): frame id like f102
  - box (array of number, optional): only with frame=: [x0,y0,x1,y1] on a 0-1000 grid of that image as you saw it (origin top-left), the region to zoom on
  - group (string, optional): segment group id like s3, object id like i1, or instance like #12
  - n (integer, optional): how many frames, 1-4 (default 4)

segment(frame, box, word, frames, within, same_as)
  Cut the thing you box. Draw box=[x0,y0,x1,y1] (0-1000 grid, origin top-left) around ONE thing on the image of frame you were shown, and name it with word. The segmenter cuts what is in the box and follows it into the extra frames (at most 5 more) where you know the same thing is visible. Returns groups s1, s2, ... (the boxed thing marked BOXED, then look-alikes found in those frames), a labeled overlay with your box in yellow, and white-background cutouts. within='s3' or '#12' restricts the cut to inside that parent (for parts).
  - frame (string, required): the frame you drew the box on, like f102 (the id stamped on the image)
  - box (array of number, required): [x0,y0,x1,y1] on a 0-1000 grid of that image as you saw it; a tight box around one thing
  - word (string, required): what you call it, 1-2 plain nouns, e.g. 'cup', 'dish rack', 'handle'. Used as the label and to help follow the same kind of thing into the other frames; the box decides what gets cut.
  - frames (array of string, optional): optional extra frames like ['f117','f131'] where the same thing is visible, to get a multi-view group (needed for a complete lift)
  - within (string, optional): optional parent: a group id 's3', an object id 'i1' or an instance id '#12'
  - same_as (string, optional): optional: a group or object id that the boxed thing IS (the same physical thing, e.g. you re-segmented it in other frames); the boxed group joins that object, saving a same() call

same(groups, into)
  Declare that several segment groups are one physical thing (the same object re-segmented in other frames, or split by the segmenter). Returns an object id i1 that unites their frames; lift(i1) votes with all of them. Add to an existing object with into='i1'. Only you can judge this: use the cutouts, the positions in the frames, and the 3D overlap that segment lists.
  - groups (array of string, required): group ids like ['s1','s10']; an object id like 'i1' may be included to merge it
  - into (string, optional): optional existing object id like 'i1' to add the groups to

lift(group, thorough)
  Turn a segment group (s3) or an object (i1, several groups declared the same thing; all their frames vote together) into a 3D instance (#id) and add it to memory. thorough=false: uses only the frames it has (fast; enough for anchors). thorough=true: gathers more views by geometry until the instance stops growing (use for the final answer). Returns size, centroid and an acceptance image with the lifted gaussians outlined in red.
  - group (string, required): segment group id like s3, or object id like i1
  - thorough (boolean, optional)

answer(ids, evidence)
  Final answer: the instance id(s) the sentence refers to (several when the sentence is plural, names a whole kind, or names several things: list every fitting instance), plus evidence: one triple per condition in the sentence, with how you know it (image frame + step).
  - ids (array of string, required): instance ids like ['#21']
  - evidence (array of object, required)
      . subject (string): an id (#5, s3, i1) or a name
      . relation (string): the relation word from the sentence, e.g. 'next to', 'closest to', 'under'
      . object (string): an id or a name
      . how (string): e.g. 'image f12 step 3'

report(status, detail, candidates)
  Stop without a confident answer. status 'ambiguous' (list the tied candidates) or 'not_found' (say what was searched and how much of the scene).
  - status (string in ambiguous/not_found, required)
  - detail (string, required)
  - candidates (array of string, optional)
\end{lstlisting}

\end{document}